\documentclass[10pt,journal,compsoc]{IEEEtran}
\usepackage{graphicx}
\usepackage{comment}
\usepackage{amsmath,amssymb} % define this before the line numbering.
\usepackage{color}

\usepackage{enumitem}
\usepackage{multicol}

\usepackage{multirow}
\usepackage{enumitem}
\usepackage{multicol}

\usepackage{multirow}
\usepackage[colorlinks,
            linkcolor=red,
            anchorcolor=blue,
            citecolor=green]{hyperref}

\usepackage{booktabs}
\usepackage{pifont}% http://ctan.org/pkg/pifont
\newcommand{\cmark}{\ding{51}}
\newcommand{\xmark}{\ding{55}}
\usepackage{url}

\ifCLASSOPTIONcompsoc
  \usepackage[nocompress]{cite}
\else
  \usepackage{cite}
\fi

\ifCLASSINFOpdf
\else
\fi

\begin{document}
\title{CRIC: A VQA Dataset for Compositional Reasoning on Vision and Commonsense}

\author{Difei~Gao,~\IEEEmembership{Student Member,~IEEE,}
        Ruiping~Wang,~\IEEEmembership{Member,~IEEE,}\\
        Shiguang~Shan,~\IEEEmembership{Senior~Member,~IEEE,}
        and~Xilin~Chen,~\IEEEmembership{Fellow,~IEEE}
\IEEEcompsocitemizethanks{
\IEEEcompsocthanksitem D. Gao, R. Wang, S. Shan, and X. Chen are with the Key Laboratory of Intelligent Information Processing of Chinese Academy of Sciences (CAS), Institute of Computing Technology, CAS, Beijing 100190, China.\protect\\
% note need leading \protect in front of \\ to get a newline within \thanks as
% \\ is fragile and will error, could use \hfil\break instead.
E-mail: difei.gao@vipl.ict.ac.cn, \{wangruiping, sgshan, xlchen\}@ict.ac.cn
}% <-this % stops an unwanted space
}

% The paper headers
\markboth{IEEE TRANSACTIONS ON XXXXX}%
{Shell \MakeLowercase{\textit{et al.}}: Bare Demo of IEEEtran.cls for Computer Society Journals}

\IEEEtitleabstractindextext{%
\begin{abstract}

Alternatively inferring on the visual facts and commonsense is fundamental for an advanced VQA system. This ability requires models to go beyond the literal understanding of commonsense. The system should not just treat objects as the entrance to query background knowledge, but fully ground commonsense to the visual world and imagine the possible relationships between objects, e.g., ``fork, can lift, food''.
To comprehensively evaluate such abilities, we propose a VQA benchmark, CRIC, which introduces new types of questions about \textbf{C}ompositional \textbf{R}easoning on v\textbf{I}sion and \textbf{C}ommonsense, and an evaluation metric integrating the correctness of answering and commonsense grounding. To collect such questions and rich additional annotations to support the metric, we also propose an automatic algorithm to generate question samples from the scene graph associated with the images and the relevant knowledge graph. We further analyze several representative types of VQA models on the CRIC dataset. Experimental results show that grounding the commonsense to the image region and joint reasoning on vision and commonsense are still challenging for current approaches. The dataset is available at \url{https://cricvqa.github.io}.

\end{abstract}

% Note that keywords are not normally used for peerreview papers.
}

\maketitle

\IEEEdisplaynontitleabstractindextext

\IEEEpeerreviewmaketitle

\begin{figure}
\centering
\includegraphics[height=8.40cm]{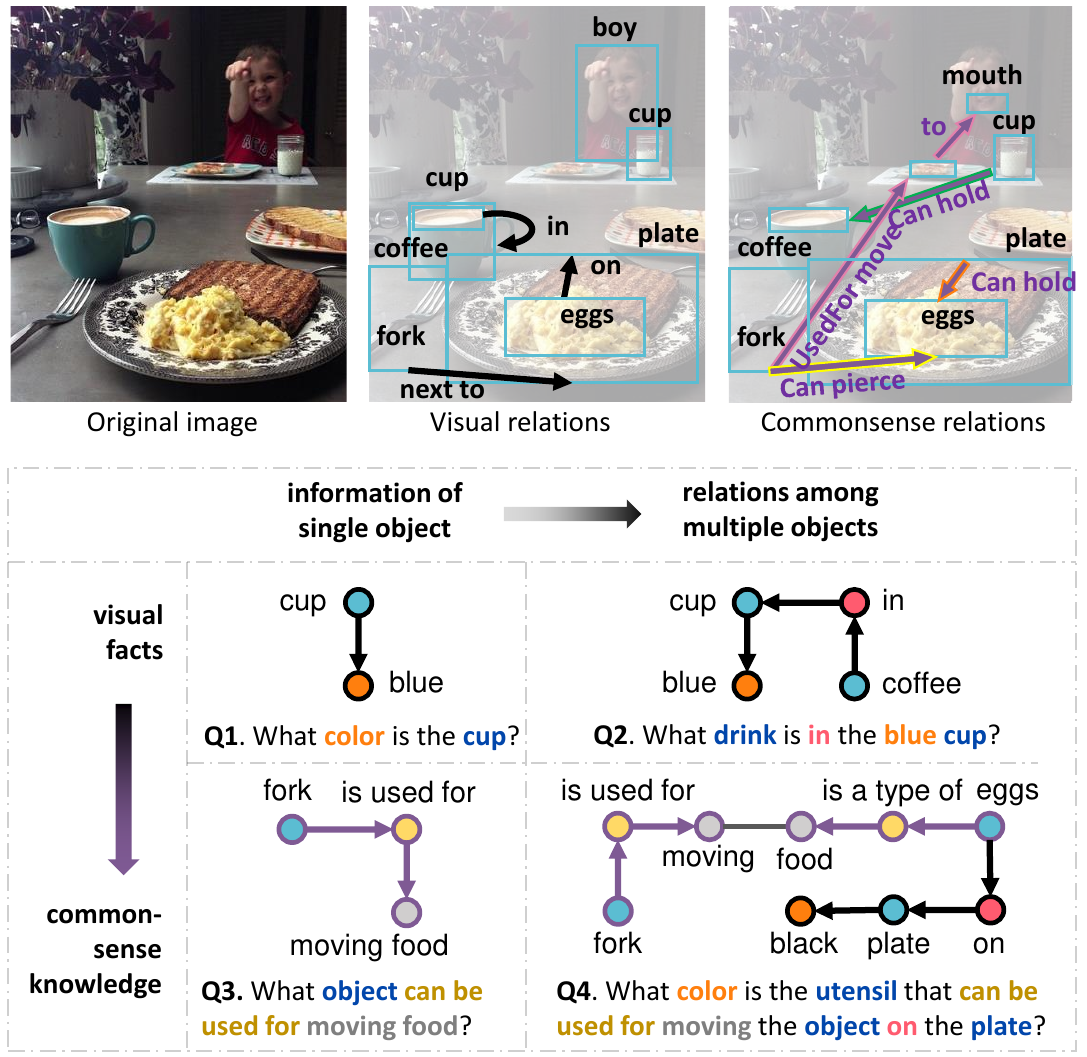}
\caption{Examples of four styles of questions. \textbf{Q1.} Querying visual information of an object. \textbf{Q2.} Multi-hop reasoning on visual relations. \textbf{Q3.} Querying non-visual knowledge about an object. \textbf{Q4.} Multi-hop reasoning on both visual and commonsense relations. Black arrow indicates the visual relation, and purple arrow indicates the commonsense relation.}
\label{fig1}
\end{figure}

\IEEEraisesectionheading{\section{Introduction}\label{sec:introduction}}
\IEEEPARstart{V}{isual} intelligence has made great progress in many specific tasks, such as image classification \cite{deng2009imagenet,krizhevsky2012imagenet,he2016deep}, object detection \cite{ren2015faster,he2017mask}, and relationship detection \cite{zhang2017visual,li2017vip}. However, it is still a formidable challenge to answer a natural language question about an image (i.e., Visual Question Answering task, VQA), which requires a system to realize a wide range of abilities.  
In the past few years, ~\cite{ren2015exploring,malinowski2014multi} first propose the VQA benchmarks, where the tasks are to answer relatively simple questions about the object name, attribute, like \textbf{Q1} in Fig.~\ref{fig1}. Further works improve the practicability and scope of the VQA task along with two orthogonal directions: 1) ~\cite{antol2015vqa,johnson2017clevr,hudson2019gqa} extend the questions about querying information of a \emph{single} visual object to questions that require multi-hop reasoning on visual relations among \emph{multiple} objects, like \textbf{Q2}. 2) ~\cite{wang2018fvqa,shah2019kvqa, cao2021knowledge} enhance the questions from only querying relatively shallow visual information of an object to querying non-visual knowledge of an object, like \textbf{Q3}. This type of questions usually require relatively simple vision abilities, such as object recognition.

However, to operate in the real world, an advanced AI agent should not only be able to reason on visual relations among multiple objects or non-visual knowledge about a single object, but to jointly infer on commonsense relations among entities and perform multi-hop reasoning on vision and knowledge. For example, to answer \textbf{Q4} in Fig.~\ref{fig1}, an AI agent is required to not only infer the explicit semantic spatial relation, \emph{eggs} on \emph{plate} based on \textbf{\emph{what it sees}} in the image, but more importantly infer the implicit commonsense relation between the objects based on \textbf{\emph{what it knows}} about the world, \emph{fork} can move \emph{eggs}. It is a higher level of visual commonsense reasoning. The AI agents should not just treat objects as the entrance to query background knowledge, but fully ground commonsense to the visual world and imagine the possible relationships between objects, as shown in the top of Fig.~\ref{fig1}.
Therefore, this paper aims to extend the VQA task along with both directions and introduces a new VQA benchmark about Compositional Reasoning on vIsion and Commonsense.

The VQA task at the intersection of vision, language and commonsense makes fairly evaluating the models challenging. The commonsense-related questions derived from natural images are inevitable to mirror some priors inherent in the real world. These priors could be the hints for a model to achieve high scores by guessing the answers, e.g., the word \emph{cut} in a question could be a hint for answering \emph{knife}. Thus, to create a commonsense VQA dataset, it is crucial to reduce the commonsense priors' impact to fairly evaluate whether the model truly understands the vision and commonsense. To achieve this goal, we introduce two essential features to the CRIC. 1) We carefully design some new types of compositional questions to force models to look at the images, e.g., query the attribute of an object that meets a commonsense requirement, like \textbf{Q4}. 2) We not only use the correctness of the final answers to evaluate the model, but also the correctness of the intermediate grounding results, i.e., the model has to correctly find the object that meets the requirement of the question. Only when both two metrics are correct, one question is considered to be correctly answered. 

To achieve these two features, we need strictly control the content of questions and collect rich annotations to support the above-mentioned metrics. The cost will be very high if the dataset is purely manually collected. Therefore, we propose a generator to automatically output question-answer pairs. Specifically, we dynamically assembles the question template from predefined template components for a given the \emph{scene graph} associated with the images and the relevant \emph{knowledge graph}. Along with the question and the answer, for each sample, we also automatically generate rich annotations to ease the difficulty of diagnosing a model, including the reasoning steps and their ground truth outputs of answering questions. 

\begin{figure}
%\centering
\includegraphics[height=3.25cm]{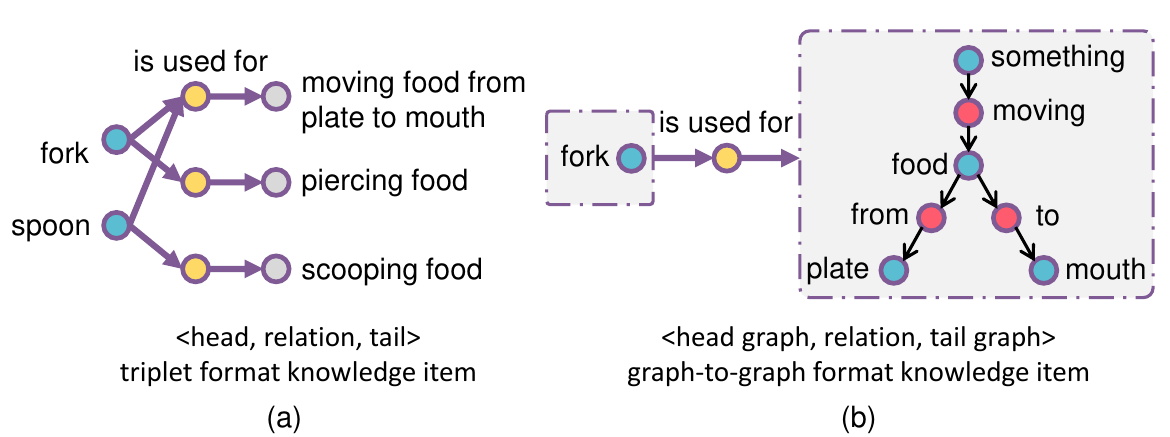}
\caption{Examples of triplet format versus graph-to-graph format knowledge items. \textbf{(a)}: The triplet format item (coming from ConceptNet~\cite{liu2004conceptnet}) usually depicts the relation between an object and an event. \textbf{(b)}: Our graph-to-graph format represents the item in a finer granularity (object-level nodes), which can be easily aligned to visual objects and depict more informative commonsense relations between objects. Here in this case, the head graph has only one single node.}
\label{fig2}
\end{figure}

To support such generator, we need to first collect scene graphs and knowledge graphs as the basis to provide object-level visual information and knowledge. Thanks to the Visual Genome dataset~\cite{krishna2017visual}, the scene graphs of images are easy to get. However, the collection of knowledge graphs faces new challenges. To generate compositional questions, we need object-level knowledge which depicts the commonsense relations between objects. However, as shown in Fig.~\ref{fig2} (a), the current format of existing available knowledge items (e.g., items in ConceptNet~\cite{liu2004conceptnet}) are on event-level, which makes commonsense relation between objects hard to be effectively represented. Rich information are simply represented in a triplet format $<$\emph{head}, \emph{relation}, \emph{tail}$>$ to represent complex semantic-level relations (e.g., \emph{is used for}, \emph{can}) between an object and an event, where the triplet has to mix up multiple objects and relations in phrase form (e.g., \emph{moving food from plate to mouth}) simply as a \emph{head} or \emph{tail} entity. Obviously, the phrase type entity is difficult to be aligned to visual objects, let alone to represent our desired commonsense relations between objects. To tackle this issue, we collect the original items from existing knowledge graph (e.g., ConceptNet), then decompose phrase-formed entities into a finer granularity, i.e., object-level nodes, and re-organize the original \emph{head} and \emph{tail} entities as graphs to obtain graph-to-graph format, as shown in Fig.~\ref{fig2} (b). The graph-to-graph format item is much easier to be aligned to objects in images and can depict more informative commonsense relations (e.g., \emph{can lift}, \emph{can move}, etc.) between the objects (e.g., \emph{fork}, \emph{can lift}, \emph{food}).  

We further evaluate several representative types of VQA models on the CRIC dataset to analyze the advantages and disadvantages of them. In addition, we leverage the well studied modular network~\cite{hu2017learning,johnson2017inferring} and our provided output of every function in programs to provide a detailed analysis of the main challenges of the CRIC. The experiments show that current joint representation of commonsense and vision limits the grounding the commonsense to the images and performing compositional reasoning on vision and commonsense, and cumulative error restricts the multi-step reasoning performance of modular networks. 

\begin{table*}[h]
\caption{Comparison of various VQA datasets. The last three columns are about the additional annotations provided by the datasets. The CRIC contains compositional questions for commonsense reasoning and provides rich annotations.}
\centering
\resizebox{\textwidth}{!}{
\begin{tabular}{lccccccc}
\toprule[1.2pt]
Dataset    & 
\multicolumn{1}{c}{\begin{tabular}[c]{@{}c@{}} Year of \\ Publication\end{tabular}} & 
\multicolumn{1}{c}{\begin{tabular}[c]{@{}c@{}}Num. of \\ Images\end{tabular}} & \multicolumn{1}{c}{\begin{tabular}[c]{@{}c@{}}Num. of \\ Questions\end{tabular}} & 
\multicolumn{1}{c}{Task Focus} & 
\multicolumn{1}{c}{\begin{tabular}[c]{@{}c@{}}Scene \\ Graph\end{tabular}} & \multicolumn{1}{c}{\begin{tabular}[c]{@{}c@{}}Knowledge \\ Graph\end{tabular}} & \multicolumn{1}{c}{\begin{tabular}[c]{@{}c@{}}Functional \\ Program\end{tabular}} \\ %& \multicolumn{1}{c}{\begin{tabular}[c]{@{}c@{}}Intermediate \\ Results\end{tabular}} \\ 
\midrule[1pt]
\textbf{CRIC (Ours)}               &    - & 96K	    & 494K  & \multicolumn{1}{c}{\begin{tabular}[c]{@{}c@{}}Commonsense (Compositional)\end{tabular}}
                                                                                &\cmark &\cmark & \cmark \\ %& \cmark \\
OK-VQA \cite{Marino_2019_CVPR}     & 2019 & 14K       & 14K     & Unstructured Knowledge                &\xmark &\xmark & \xmark \\
KVQA \cite{shah2019kvqa}           & 2019 & 24K       & 183K    & Name Entities related Knowledge       &\xmark &\cmark & \xmark \\ %& \xmark \\
VCR \cite{zellers2018recognition}  & 2018 & 110K      & 290K    & Commonsense                           &\cmark &\xmark & \xmark \\ % & \xmark \\
FVQA \cite{wang2018fvqa}           & 2018 & 2.2K      & 5.8K    & Commonsense                           &\xmark &\cmark & \xmark \\ %& \xmark \\
KB-VQA \cite{wang2017explicit}     & 2017 & 0.7K      & 2.4K    & Commonsense                           &\xmark &\xmark & \xmark \\ %& \xmark \\
GQA \cite{hudson2019gqa}           & 2019 & 113K      & 22M     & Vision (Compositional)                &\cmark &\xmark & \cmark \\ %& \xmark \\
CLEVR \cite{johnson2017clevr}      & 2017 & 100K      & 999K    & Vision (Compositional)                &\cmark &\xmark & \cmark \\ %& \xmark \\
PQA  \cite{qi2021pqa}     & 2021 & 32.9K      & 157K    & Perceptual Reasoning                          &\xmark &\xmark & \xmark \\ % & \xmark \\
%TallyQA \cite{acharya2018tallyqa}  & 165K      & 287K    & Counting                              &\xmark &\xmark & \xmark \\ %& \xmark \\
VQA  $v$2 \cite{goyal2016making}   & 2016 & 204K      &1.1M     & Vision                                &\xmark &\xmark & \xmark \\ %& \xmark \\
VQA $v$1 \cite{antol2015vqa}       & 2015 & 204K      & 614K    & Vision                                &\xmark &\xmark & \xmark \\ %& \xmark \\
VQA-abstract  \cite{antol2015vqa}  & 2015 & 50K  & 150K    & Vision (Scene Graph)                  &\cmark &\xmark & \xmark \\ %& \xmark \\
COCO-QA \cite{ren2015exploring}    & 2015 & 69K       & 117K    & Vision                                &\xmark &\xmark & \xmark \\ %& \xmark \\
DAQUAR  \cite{malinowski2014multi} & 2014 & 1.4K      & 12K     & Vision                                &\xmark &\xmark & \xmark \\ %& \xmark \\
\bottomrule[1.2pt]
\end{tabular}
}
\label{tab_datasets}
\end{table*}

To summarize, the contributions of this paper are as follows:
1) We propose a new benchmark CRIC which introduces new types of questions for fairly evaluating the ability of compositional reasoning on visual and commonsense. 2) To build a dataset to support such task at a proper cost, we propose an dynamic template assembly dataset construction method to collect question-answer pairs and rich additional annotations. 3) To collect satisfactory knowledge items to generate compositional questions, we introduce a new graph-to-graph format for representing the knowledge items that is better in aligning entities in KG to objects in images and depicting the commonsense relations between objects. 4) Further experiments provide detailed analyses about the representation and reasoning abilities of the existing several representative types of VQA methods and the challenges of sub-tasks in CRIC.

\section{Related Work}

\textbf{VQA Dataset.}
At the early stage, COCO-QA~\cite{ren2015exploring} and DAQUAR~\cite{malinowski2014multi} focus on evaluating a range of visual abilities about querying the visual information of a visual entity, e.g., recognizing the category or attributes of an object. Further works~\cite{zhu2016visual7w,kafle2017analysis,antol2015vqa,yu2015visual}, such as VQA~\cite{antol2015vqa}, extend the scope of questions by requiring VQA systems understanding visual relations between objects. In addition, CLEVR \cite{johnson2017clevr} emphasizes the importance of a VQA system on compositional reasoning on spatial relations and provides compositional questions about synthetic images. More recently, GQA \cite{hudson2019gqa} introduces a real-image VQA dataset with compositional visual questions and a more balanced answer distribution.

Another trend in recent studies~\cite{zellers2018recognition,shah2019kvqa,wang2018fvqa,wang2017explicit,Kembhavi2017tqa,Marino_2019_CVPR} is to expand the scope of the questions in another direction by requiring some commonsense-related abilities. \cite{wang2018fvqa} introduces the FVQA dataset, of which each question relates to one knowledge triplet in Knowledge Graph. The FVQA aims to evaluate the ability of VQA systems on understanding non-visual background knowledge of the objects. \cite{shah2019kvqa} introduces KVQA dataset containing questions about querying the background knowledge extracted from Wikipedia of name entities, rather than common objects, e.g., \emph{Who is to the left of Barack Obama}. The OK-VQA~\cite{Marino_2019_CVPR} provides commonsense questions that require the model to mine the background knowledge of objects from outside natural language documents, rather than Knowledge Graph. The VCR dataset \cite{zellers2018recognition} focuses on challenging commonsense reasoning questions, e.g., inferring why something happened or the mental state of a person. Unlike previous datasets, which provide external knowledge sources, VCR requires understanding the knowledge about causal relations, social interactions and physics acquiring from training samples. 

Compared to previous datasets, our proposed questions require the systems to not only use objects as the entrance to query background knowledge, but fully ground commonsense to the visual world and imagine the possible relationships of objects. We believe this is a crucial and fundamental ability for future AI agents. In Table~\ref{tab_datasets}, we display the basic statistics and main characteristics of major VQA datasets and our proposed CRIC dataset. And, a discussion about CRIC and the most relevant datasets is in Sec.~\ref{discussion}.

\textbf{Knowledge Graph.}
Structured Knowledge Graphs (KGs) are great sources to provide explicit and well-organized information for AI agents. There are two types of KGs widely used in AI researches, that is, world knowledge KGs, such as DBpedia~\cite{auer2007dbpedia}, Freebase~\cite{bollacker2008freebase}, and commonsense KGs, such as ConceptNet~\cite{liu2004conceptnet} and Webchild~\cite{tandon2014webchild}. The world knowledge KGs are broadly used in NLP and AI communities~\cite{berant2013semantic,weston2014memory,sukhbaatar2015end,yang2015wikiqa} for supporting the AI agent in answering knowledge related question. In recent years, many inspiring computer vision works~\cite{wang2018fvqa,Gu_2019_CVPR,narasimhan2018out,Qi_2019_CVPR,Aakur2019GeneratingOW,narasimhan2018straight,li2017incorporating, zareian2020bridging} attempt to introduce commonsense KGs in their tasks and use external knowledge to expand the capability of visual systems. 

In commonsense KGs, knowledge is typically represented by a large set of items in triplet format $<$\emph{head}, \emph{relation}, \emph{tail}$>$, where \emph{head} and \emph{tail} are two entities in the KG and \emph{relation} indicates the relationship between them. Compared with world knowledge KGs, the items in commonsense KGs have one unique characteristics: a large number of entities are informative phrases depicting an event rather than a real ``entity'', e.g., $<$\emph{bus}, \emph{is used for}, \emph{transporting students to school}$>$. The rich information depicting the relationships between the objects is simply in a phrase, e.g., \emph{bus} transport \emph{students}, \emph{bus} move to \emph{school}. Therefore, compared to previous works that directly use these KGs into their tasks, we furthur decompose the triplet items into a new format to mine the information hidden in the entities and evaluate the VQA systems on understanding such more complicated knowledge.   

\textbf{Dataset Construction.} 
Many existing VQA datasets~\cite{malinowski2014multi,antol2015vqa,goyal2016making,Marino_2019_CVPR,wang2018fvqa} collect free-form and open-ended visual questions from human annotators. Another branch of works~\cite{ren2015exploring,kafle2017analysis,johnson2017clevr,hudson2019gqa} which focus on evaluating some specific abilities of VQA models, propose to generate questions by the template-based method automatically. \cite{ren2015exploring} designs rules to convert image descriptions into some predefined types of questions. \cite{johnson2017clevr} proposes to generate compositional questions of synthetic images by filling predefined question templates with elements in scene graphs. \cite{kafle2017analysis,hudson2019gqa} design more diverse templates to generate rich questions querying about natural images. 

Previous automatic question generation methods usually require predefining almost all possible templates and are thus less efficient and scalable for generating our desired questions, which involve large concept vocabulary and commonsense knowledge. To address this problem, we propose a new question generator to dynamically assemble the question template from predefined basic template components given the scene graph of an image and a collected knowledge graph. 

\begin{figure*}
\centering
\includegraphics[height=7.6cm]{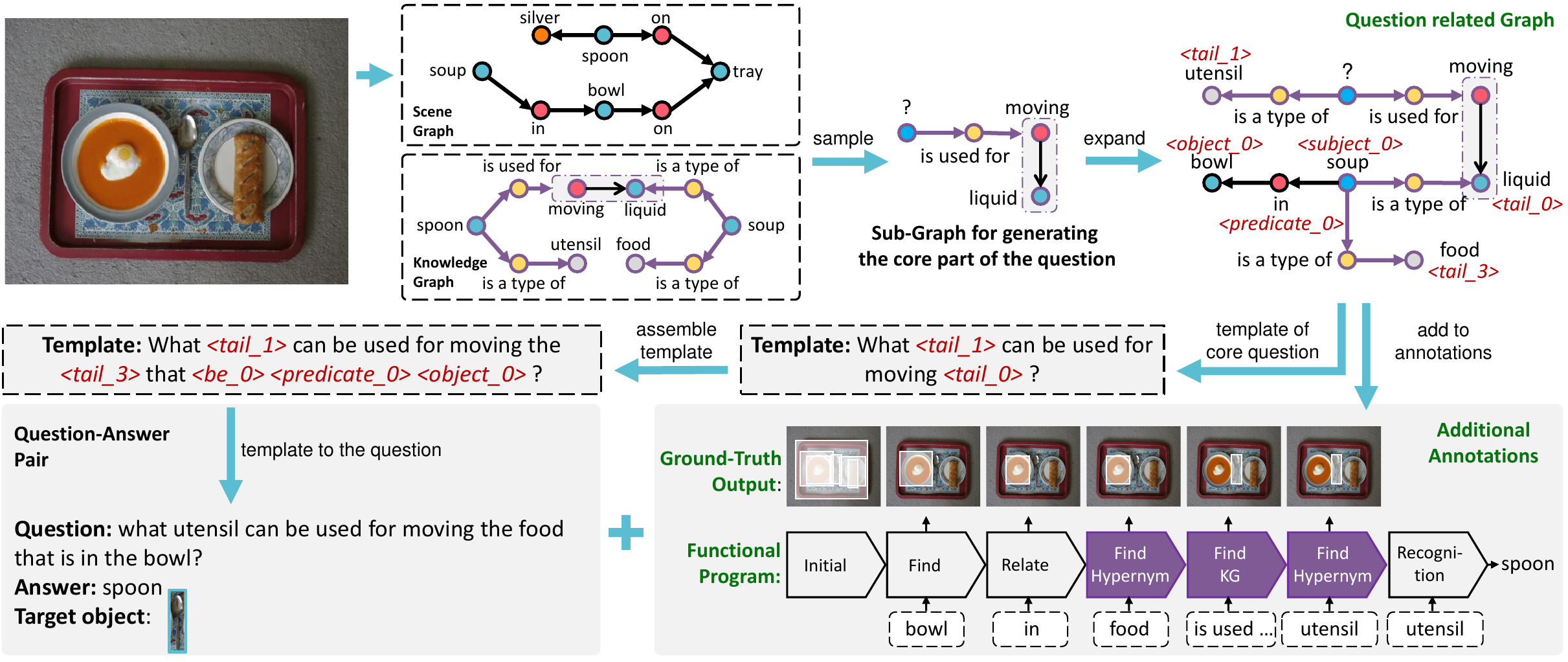}
\caption{Overview of the QA sample generation process. The question and corresponding annotations are automatically generated from the Scene Graph of a given image and the Knowledge Graph. Our method first selects the proper part of the Scene Graph and the Knowledge Graph that can generate a question, then assembles the question template from predefined template components, and finally generates the question-answer pair along with rich annotations. }

\label{data_collect}
\end{figure*}

\section{Dataset Construction}
\label{dataset_construction}
\textbf{Overview.} 
We introduce CRIC, a new task that challenges AI systems to ground the commonsense into the visual world and perform multi-hop reasoning on the real image and the knowledge graph. To evaluate such abilities, the CRIC provides 494K balanced and compositional questions on 96K images along with 3.4K knowledge items. Besides, to ease the difficulty of designing robust and interpretable systems, the dataset collects rich annotations about task decomposition, scene graph and knowledge graph related to every question, reasoning steps, and corresponding results for answering every question.
% We present VCR, a new task that challenges vision systems to holistically and cognitively understand the content of an image
Our dataset is constructed in five main steps: 1) process scene graphs, 2) collect and parse knowledge triplets, 3) define basic functions that the question will involve, 4) automatically generate QA samples from the scene graphs and the knowledge graph, and 5) obtain additional annotations, as shown in Fig.~\ref{data_collect}.

\textbf{Scene Graph Processing.}
The CRIC dataset utilizes the 108K images of Visual Genome and their corresponding Scene Graph annotations to generate QA samples. The scene graph is a structured representation of an image, where nodes are objects annotated with attributes and edges connect two related objects. In this stage, we first clean up the scene graphs by filtering rare concepts and merging synonyms. Our processed scene graphs contain 1291 distinct objects, 267 attributes, and 210 relationships. It is also observed that one object in the image might correspond to multiple object IDs and bounding boxes in the scene graph. This will cause ambiguity in the later question generation procedure. Thus, we merge bounding boxes that correspond to the same object name and have a high IoU ($>$ 0.7).  

\begin{figure}[t]
%\centering
\includegraphics[height=13.0cm]{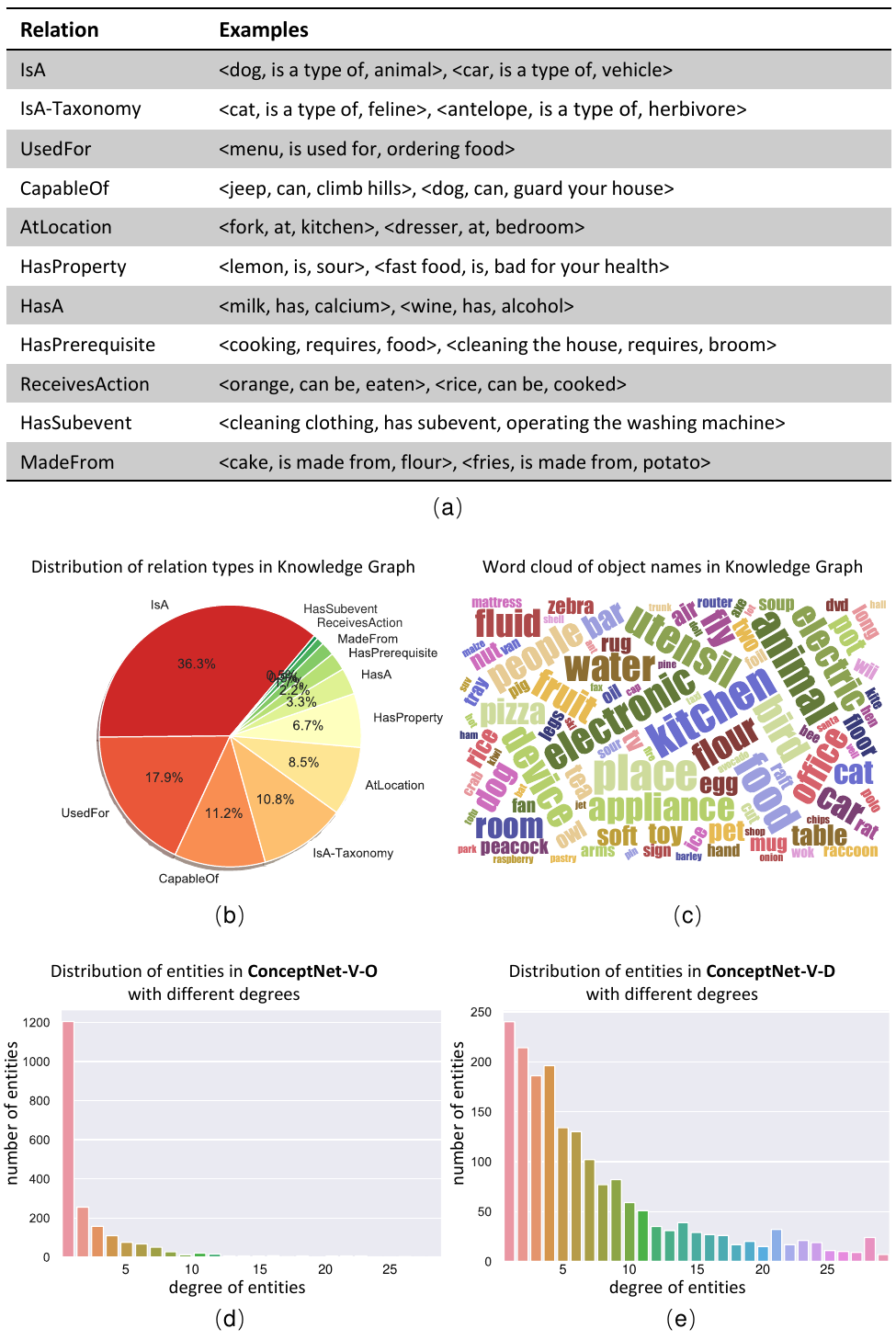}
\caption{Statistics of collected knowledge items. \textbf{(a)}: Relation types and their examples in our selected knowledge items. \textbf{(b)}: The distribution of relation types in our selected knowledge graph. \textbf{(c)}: Word cloud for frequent object names in our knowledge graph. \textbf{(d)} and \textbf{(e)}: The distribution of entities in the original KG and processed KG with different degrees. Processing KG digs out the hidden relationships between the items and dramatically increases the density of the KG.}
\label{statistic_kg}
\end{figure}

\begin{figure*}[thp]
\centering
\includegraphics[height=3.5cm]{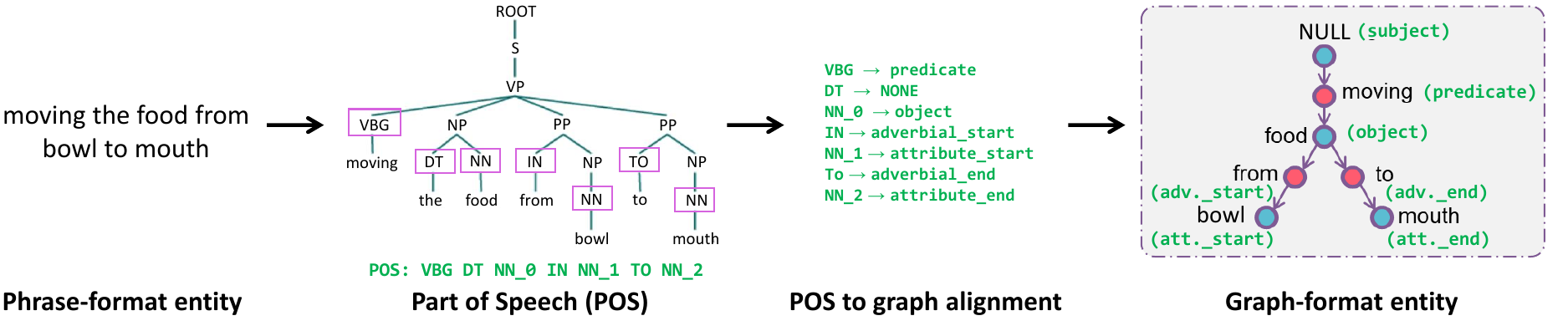}
\caption{Pipeline of parsing a phrase-format entity into a graph-format entity. The ``NONE'' in \textbf{POS to graph alignment} indicates that the word doesn't need to be assigned to the graph template. The ``NULL'' in \textbf{Graph-format entity} indicates that no word is assigned to this element in the template.}
\label{kb_parser}
\end{figure*}

\textbf{Knowledge Graph Processing.} 
The purposes of this stage are 1) collecting commonsense knowledge that is useful in daily life and related to the images in Visual Genome and 2) parsing the triplet format of knowledge items into the graph-to-graph format. 

In this paper, our knowledge graph is extracted from a large-scale commonsense Knowledge Graph, ConceptNet~\cite{liu2004conceptnet}. The knowledge in ConceptNet is collected from a variety of resources, such as crowd-sourced resources (e.g., Open Mind Common Sense \cite{singh2002open}) and expert-created resources (e.g., WordNet \cite{miller1995wordnet} and JMDict \cite{breen2004jmdict}), and is represented in triplet format $<$\emph{head}, \emph{relation}, \emph{tail}$>$. To collect satisfactory knowledge triplets, we first query the ConceptNet with all the concepts in the processed scene graphs and obtain about 225K triplets. Then, we filter the triplets with relations that are unnatural to appear in a vision-related question, e.g., $<$\emph{person}, \emph{Desires}, \emph{own a house}$>$. In addition, we carefully refine the items to keep similar events expressed in the same style (e.g., same sentence structure and predicate) to avoid some special words being hints for guessing latter generated questions based on these items. Besides, the items about one thing in ConceptNet sometimes are incomplete, e.g.,  ConceptNet tries to record which objects contain calcium, while only \emph{milk} and \emph{ice cream} are recorded. This could cause our generator to output ambiguous questions or incorrect answers. Thus, we further collect 372 knowledge items from Wikipedia to make the knowledge in ConceptNet more complete in such cases. We also collect two types of categorization knowledge of the objects from Wikipedia and WordNet, where one type is about trivial category knowledge, e.g., $<$\emph{cat}, \emph{IsA}, \emph{animal}$>$, which is used for referring object in question, e.g., ``which animal can ...?''; another type is about more professional taxonomy, e.g., $<$\emph{cat}, \emph{IsA}, \emph{feline}$>$, which is used for querying models, e.g., ``which animal is a feline?''. Finally, we obtain 3,439 carefully selected knowledge triplets with 11 types of relations, e.g., \emph{IsA}, \emph{UsedFor}, \emph{HasA}. 

In Fig.~\ref{statistic_kg} (a), we present the selected 11 types of relations and show some examples of each type. The distribution of relation types in the knowledge graph is shown in Fig.~\ref{statistic_kg} (b). We can see that our collected knowledge items are roughly evenly distributed over the relation types of \emph{UsedFor}, \emph{CapableOf}, \emph{IsA-Taxonomy}, \emph{AtLocation}, \emph{HasProperty}. In addition, the word cloud for frequent object names appearing in the knowledge items is shown in Fig.~\ref{statistic_kg} (c). We can see that the knowledge items are mainly about objects commonly used or seen in daily life, such as kitchen utensils, foods, appliances, and animals.

In the following, we parse the selected knowledge triplets into the graph-to-graph format, as shown in Fig.~\ref{fig2} (b). This goal is achieved by developing a simple rule-based phrase-to-graph parser. Our parser first utilizes the Stanford CoreNLP~\cite{manning-EtAl:2014:P14-5} to obtain POS tags of a phrase. Then, we build a set of POS-to-graph mapping templates, where the phrase-format entity can be automatically mapped to a graph according to the POS tags of the words, as shown in Fig.~\ref{kb_parser}. In Fig.~\ref{statistic_kg} (d) and (e), we compare the ``degree'' of entities (head and tail entities) in our processed KG (denoted as ConceptNet-V-D, where ``-V'' denotes ``vision'', and ``-D'' denotes ``Decomposed'') with the original triplet format KG (denoted as ConceptNet-V-O). For triplet format KG, since the basic elements in them are entities, we consider two entities as connected (contribute 1 degree) when they are exactly the same; while in graph-to-graph format KG, two entities (a.k.a graph-format entities) are considered as connected when they share at least one object node, since the basic element in the new format is the node in a graph-format entity. We can find that introducing the new format of item digs out hidden relations between the items in the original KG and dramatically increases the density of the Knowledge Graph. The dense connected KG can facilitate us in later building challenging multi-hop reasoning questions.

\begin{figure}[tph]
\includegraphics[height=10cm]{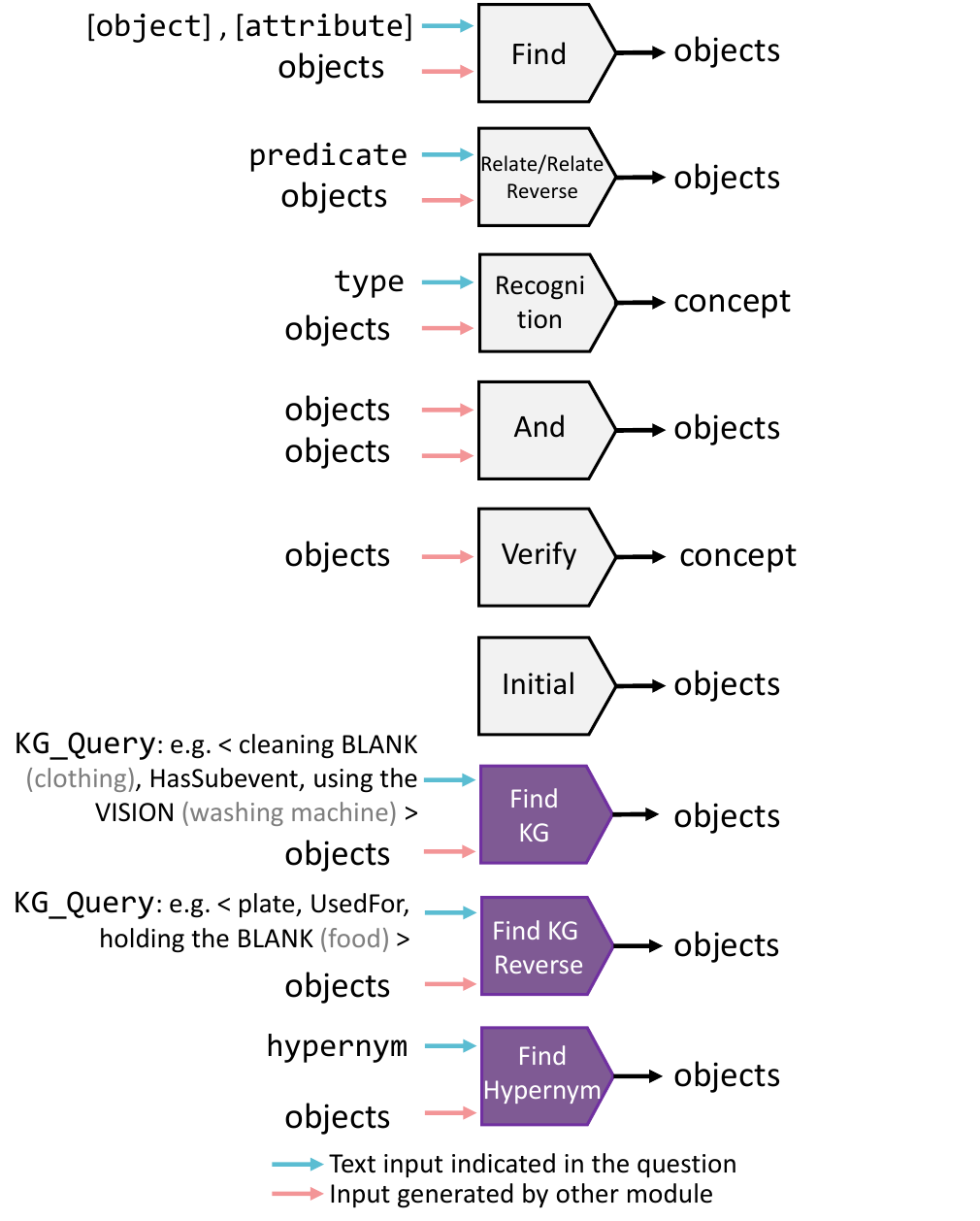}
\caption{Catalog of basic functions evaluated in questions of the CRIC dataset.}

\label{all_task}
\end{figure}

% The illustration of visual basic functions is in Appendix~\ref{app_fun}

\textbf{Function Definition.}
At this stage, we define the basic functions in the CRIC about compositional reasoning on vision and commonsense. There are ten basic functions in the CRIC dataset, where three commonsense related functions are unique in CRIC, and the others are similar to those in CLEVR and GQA, as shown in Fig.~\ref{all_task}. 

Specifically, two functions relate to basic logical operations: ``And'', ``Verify''. Four functions are about basic abilities of reasoning on the image: ``Find'', ``Relate'', ``Relate Reverse'', ``Recognition'', where ``Find'' indicates find the object for an given object or attribute name, ``Relate'' indicates the task that given \emph{subject} and \emph{predicate} in a scene graph relationship $<$\emph{subject}, \emph{predicate}, \emph{object}$>$, the model needs to locate the region of \emph{object}, while ``Relate Reverse'' indicates that given \emph{predicate} and \emph{object}, the model locates the region of \emph{subject}, and ``Recognition'' corresponds to a set of subtasks, such as, recognizing color, recognizing animal, etc. Moreover, we propose three new functions related to the commonsense reasoning: ``Find KG'', ``Find KG Reverse'', ``Find Hypyernym''. ``Find KG'' and ``Find KG Reverse'' require the model to find image regions that satisfy a commonsense query, e.g., find a proper object and fill it in the BLANK in query $<$\emph{cleaning} BLANK, \emph{HasSubevent}, \emph{using the} VISION \emph{(washing machine)}$>$ to make the statement in accordance with commonsense. Note that, the commonsense query could be a multi-modal query, e.g., the \emph{washing machine} in the above query can be a text or a region containing \emph{washing machine} which is outputted by another module. ``Find Hypyernym'' is required to find the object for a given category name, e.g., find all objects in an image which belong to \emph{vehicle}. Finally, we design a simple function, ``Initial'' to attend on all objects, which is usually used at the functional program's start. In the appendix, we provide more details of these functions.

\textbf{Template Collection \& Question Generation.}
In this section, we introduce a scalable and low-cost question generator to automatically create numerous questions by imitating the procedure of humans creating a complex question. 
As shown in Fig.~\ref{data_collect}, one question is generated from a dynamically composed question template based on a sub-graph composed of a sub-scene graph and a sub-knowledge graph. 
To achieve this goal, we first need to build two types of template components. One type is to \textbf{query} one element in a visual triplet or a knowledge item, e.g., the template of querying color ``what color/which color/... is the $<$\emph{subject}$>$?'', where $<$\emph{subject}$>$ will be filled in based on the graph annotation. 
The other one is about how to use one visual triplet or one knowledge item to \textbf{decorate} one object, e.g., the $<$\emph{object\_1}$>$ (apple) that $<$\emph{be\_2}$>$ (is) $<$\emph{predicate\_2}$>$ (on) $<$\emph{object\_2}$>$ (the plate). To increase the diversity of the question, one template component usually has multiple versions that will be randomly chosen to generate the question. 

\begin{figure}[t]
%\centering
\includegraphics[height=11.0cm]{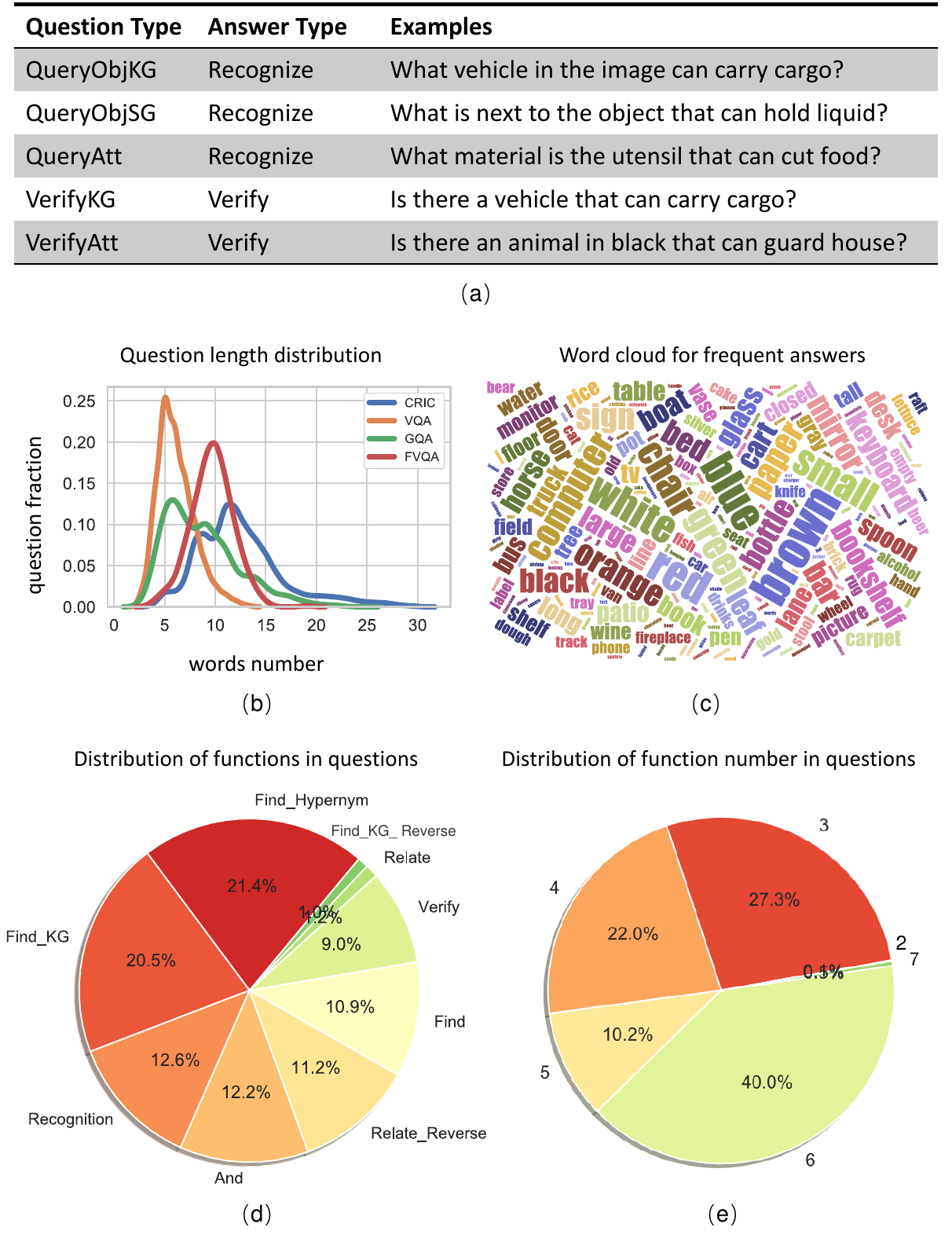}
\caption{Statistics of questions and answers in CRIC. \textbf{(a)}: 5 main categories of the CRIC questions and their examples. \textbf{(b)}: Comparison of question length distributions for different VQA datasets. The questions in the CRIC are generally much longer and have a wide range of lengths. \textbf{(c)}: Word cloud for
frequent answers. \textbf{(d) \& (e)}: Distributions of functions in overall questions and function number of programs in overall questions. }
\label{statistic_qa}
\end{figure}

\begin{figure*}[t]
\centering
\includegraphics[height=7.6cm]{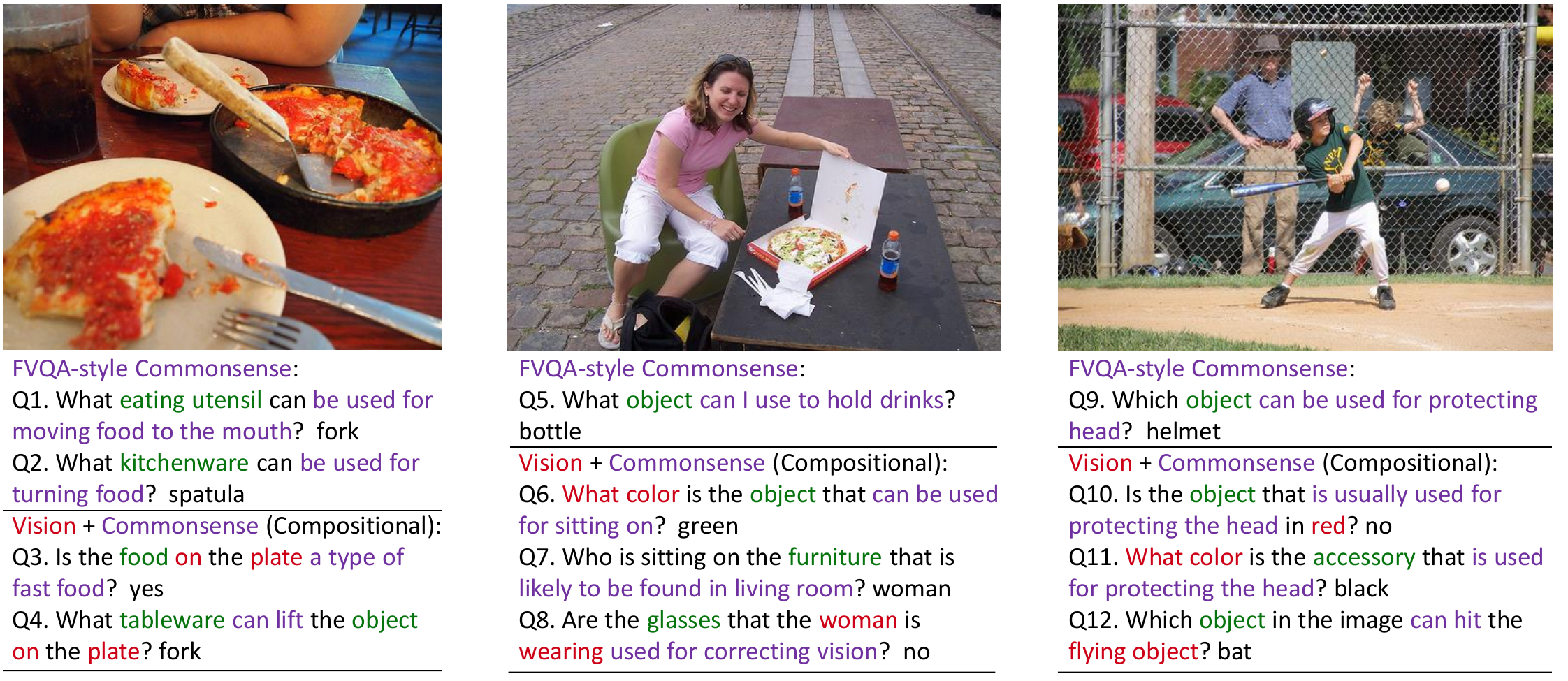}
\caption{Some example questions from the CRIC dataset. Our dataset contains not only the relatively simple commonsense question type similar to FVQA~\cite{wang2018fvqa}, but also the proposed unique compositional questions for reasoning on vision and commonsense knowledge. These compositional questions can avoid models answering by guessing through requiring recognizing visual attributes or spatial reasoning after commonsense reasoning to force models looking at the image to answer.}
\label{fig_examples}
\end{figure*}

Then, the question is generated in the following steps: 1) One visual relation or knowledge item is selected to generate the core part of the question. 2) Proper relations and attributes are added to decorate the core question, when the core question contains limited information to precisely locate the image region, or we want to provide additional information to locate the image region better. 3) The template of the question will be automatically composed of basic template components. 4) The blanks in the template will be filled in based on the scene graph and the knowledge graph. In Fig.~\ref{fig_examples} and further in the appendix, we show some QA samples in the CRIC. 

%\textbf{New Question Types for Reducing Language Priors in Samples.}
Note that, the questions derived from natural images and common knowledge will naturally mirror some priors in the real world. These priors could be the hints for models to achieve high scores by guessing the answers without truly understanding the images and knowledge, e.g., the word \emph{cut} in a question could be a hint for answering \emph{knife}. Therefore, to propose questions that can fairly evaluate the models, we carefully design some \textbf{new types of compositional commonsense questions to avoid models answering by guessing} (the question types in the CRIC are shown in Fig.~\ref{statistic_qa} (a)). For example, we design {\em QueryAtt}, {\em QueryObjSG} and {\em VerifyAtt} types of questions to require recognizing visual attribute or spatial reasoning after commonsense reasoning to force models looking at the image to answer, e.g., Q6, 7, 10, etc. in Fig.~\ref{fig_examples}. Besides, we let the question generator sometimes replace an element in knowledge items with a referring expression, e.g., Q4 in Fig.~\ref{fig1} and Q4, 12 in Fig.~\ref{fig_examples}, to avoid frequently involved knowledge expressions being used as hints.

\textbf{Obtaining Additional Annotations.}
For every QA sample in the CRIC dataset, we provide the question and answer, along with additional annotations, including the sub-scene graph and sub-knowledge graph used in answering the question, the representation of the question in the form of a functional program and the ground-truth output of every function in the program, as shown in Fig.~\ref{data_collect}. The sub-scene graph \& sub-knowledge graph and the functional program can be automatically generated during the question generation. To collect the ground-truth of each function, at every step of the program, we search on the scene graph and knowledge graph to find candidate objects that satisfy all previous functions' requirements simultaneously.
%Note that, our functional programs are already written in Reverse Polish Notation \cite{burks1954analysis}, 

Now, we obtain 3M automatically generated QA samples. However, these samples are highly unbalanced. To avoid the model overfitting on the bias of the dataset, we balance the dataset by filtering the QA samples based on the distribution of answers and the distribution of knowledge items involved in the questions. Finally, we obtain the CRIC dataset which contains \textbf{96,241 images} with \textbf{494,350 QA samples} and \textbf{3,439 knowledge triplets}. To guarantee the quality of the dataset, it is crucial for validating the correctness of the annotations. We have manually validated the correctness during dataset construction to ensure quality at each key stage. {\em 1) Module validation}: Our question and additional annotation generators are both modularized. We test on more than 100 samples in different images for each module to confirm them accurately work. {\em 2) Consistency validation}: A natural language question and corresponding function layout are two expressions for the same question. Thus, we validate the correctness of both by checking if the two annotations are mutual-consistent, e.g., if their answers are the same. {\em 3) Final results validation}: After generating, we then randomly select about 10,000 samples (i.e., about 2\% out of the total 494,350 questions) from images covering various scenes to check their correctness.

The images and corresponding QA samples of the dataset are randomly split into the train (70\%), validation (10\%), and test (20\%). The question contains on average 12 words and involves on average 6 functions.  In Fig.~\ref{statistic_qa} (b), we show the distribution of the question length of several VQA datasets. The CRIC has a relatively balanced distribution of question length and is relatively longer than other datasets. In Fig.~\ref{statistic_qa} (c), from the presented frequent answers of the CRIC, we can see that although all questions are related to commonsense, a lot of answers do not come from the knowledge items, e.g., \emph{brown}, \emph{large}. These questions will force models to look at the images based on the results of commonsense reasoning. From Fig.~\ref{statistic_qa} (d), we can see that our CRIC has a relatively balanced distribution of functions, which indicates that the CRIC provides plenty of QA samples for training each sub-task. Besides, from Fig.~\ref{statistic_qa} (e), it shows that our auto-generated questions cover a wide range of complexities (the questions involve from 2 functions to 7 functions), and more than 70\% questions, which involve more than 4 functions, are relatively complex and require multi-hop reasoning.

\textbf{Evaluation Protocol.}
For evaluation, to further avoid models achieving high scores by guessing and fairly evaluate the performances of models, our evaluation metrics consider both the correctness of 1) the \textbf{answer} and 2) the \textbf{grounding results}. More specifically, for each QA sample, a VQA system is required to provide two results: the answer and one object selected from our provided candidate objects. Note that, for yes-no questions, if the answer is \emph{yes}, the model should point out the eligible object; if the answer is \emph{no}, the model should point out no object. A question is considered to be correctly answered when the two results are both correct. To better evaluate and diagnose the performance of the reasoning abilities, especially for grounding-related functions, we provide the bottom-up features \cite{anderson2017bottom} cropped by ground-truth bounding boxes as the image features. In addition, we classify the questions into two groups, Verify and Recognize, by checking if the answer is ``yes'' or ``no''.

\section{Experiments}
In this section, we evaluate the performances of four types of representative methods on the CRIC (Sec.~\ref{exp_overall}), including classical monolithic VQA models, modular VQA models, KB-aware VQA models, and recent popular visual BERT to analyze the main challenges of CRIC.

\subsection{Baselines}
\label{exp_baselines}

\hspace{\parindent}

%\noindent
\textbf{Q-Only:} Q-Only model only takes the LSTM question features as input. 

%\noindent
\textbf{I-Only:} I-Only model only takes the image feature as input. 

%\noindent
\textbf{SAN:} Stacked Attention Network~\cite{yang2016stacked} is a classical monolithic VQA model on the VQA dataset which performs two-step soft attention on the image features.

%\noindent
\textbf{Bottom-Up:} Bottom-Up~\cite{anderson2017bottom} is a classical monolithic VQA model which proposes object-level reasoning and implements soft-attention on object regions. The attended image features and question features are combined to generate the final answer.

%\noindent
\textbf{Bottom-Up+$l_{att}$:} Compared to Bottom-Up, this baseline adds a cross-entropy loss on attention score to supervise the model to attend on the correct region.

%\noindent
\textbf{MAC:} MAC~\cite{hudson2018compositional} is a state-of-the-art modular VQA model for CLEVR and GQA which decomposes a question into a series of attention-based reasoning steps.

\begin{figure}[t]
\centering
\includegraphics[width=8.7cm]{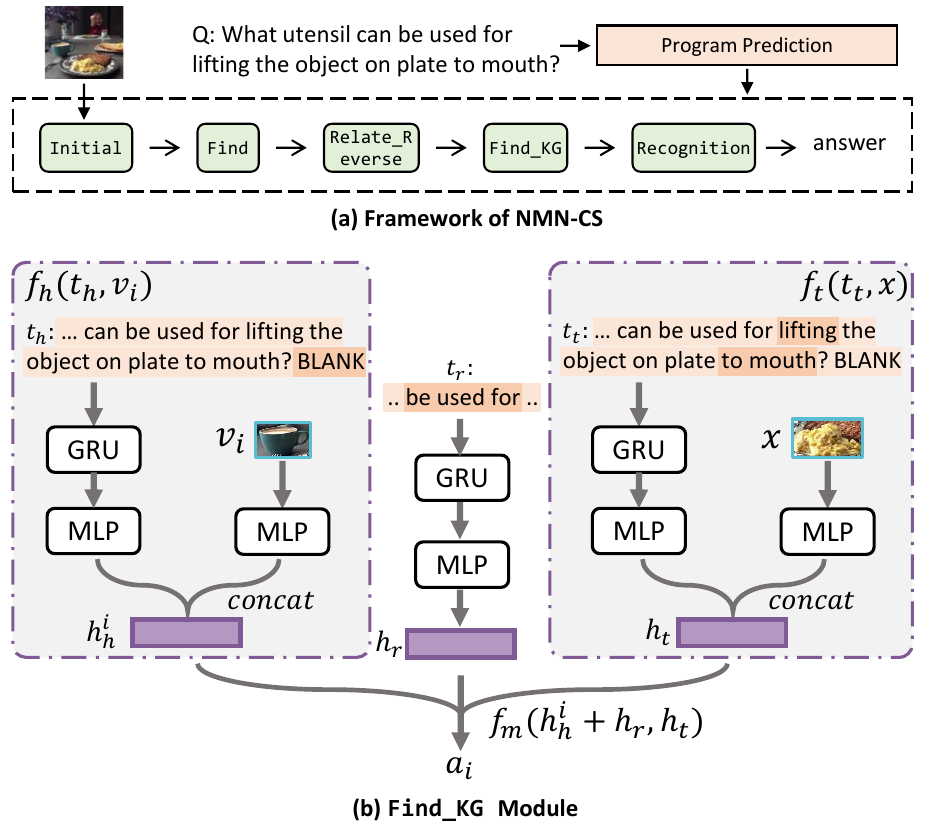}
\caption{Illustration of NMN-CS. (a) Framework of NMN-CS. A program prediction module generates the function layout, then predicted functions are performed to output the answer. (b) Find\_KG module. Three networks extract the features of three parts of the knowledge item. Then, the Find\_KG module calculates their matching score as the attention value. We add an element BLANK at the end of the question for the case that network $f_h$ doesn't need to attend on any word in the question.}
\label{find_kg}
\end{figure}

\textbf{NMN-CS:} Neural Modular Network (NMN)~\cite{hu2017learning,andreas2016neural,johnson2017inferring} is another type of state-of-the-art modular VQA model on CLEVR dataset. However, its original versions cannot directly transfer to commonsense questions, so we add some visual commonsense reasoning modules to allow the whole model to reason on the CommonSense, denoted as NMN-CS. 

More specifically, the NMN-CS builds upon neural modular networks in~\cite{hu2017learning} which contains a sequence-to-sequence network (i.e., program prediction module). It generates the function layout (a sequence of function names) and the text inputs of each function (an attended question over words) for a given question, as shown in Figure~\ref{find_kg} (a). In addition, NMN-CS contains a set of modules corresponding to each function in CRIC. The architecture of pure visual modules, such as \texttt{Find}, \texttt{Relate}, are similar to the corresponding modules in N2NMN~\cite{hu2017learning}. For commonsense-related functions, we design some simple networks to achieve these functions. 

First, for the \texttt{Find\_KG} module, as shown in Fig.~\ref{find_kg} (b), the goal is to generate the attention score $a$ over all object candidates $\{v_1, ..., v_n\}$ for a given object feature $x$ generated by another module and query sequence $t$. The query sequence contains three sub-sequences $t = [t_h, t_r, t_t]$ which indicate the head, relation, and tail in a knowledge item respectively. These sub-sequences are obtained by implementing self-attention on the given text inputs generated by program prediction network. \texttt{Find\_KG} module first separately encodes the head, relation, and tail, where head feature $h_h=[h_h^1, ..., h_h^o]$ is the combination of each $v_i$ and GRU features encoding the word sequence $t_h$, relation feature $h_r$ is encoded by text embedding layer, and tail feature $h_t$ is the combination of the visual feature generated by the previous module and the GRU feature encoding $t_t$. Then, an MLP $f_m(h_h^i + h_r, h_t)$ outputs attention score on object $v_i$ by calculating the matching score of current head, relation and tail. \texttt{Find\_KG\_Reverse} is similar to the \texttt{Find\_KG} module, where the only difference is that the positions of $x$ and $v_i$ are exchanged. In addition, the architecture of \texttt{Find\_Hypernym} is the same as the Find module, which uses the attended question feature to retrieve the objects in the image. 

The whole NMN-CS model is trained in two stages: training the program prediction module and training the neural modules. For training the program prediction module, we minimize the cross-entropy losses of predicted function names. For training the neural modules, we use the predicted function layouts to assemble the neural modules and minimize the binary cross-entropy losses for the answers. More details about this baseline are shown in the appendix.

\begin{figure}[t]
\centering
\includegraphics[height=4.7cm]{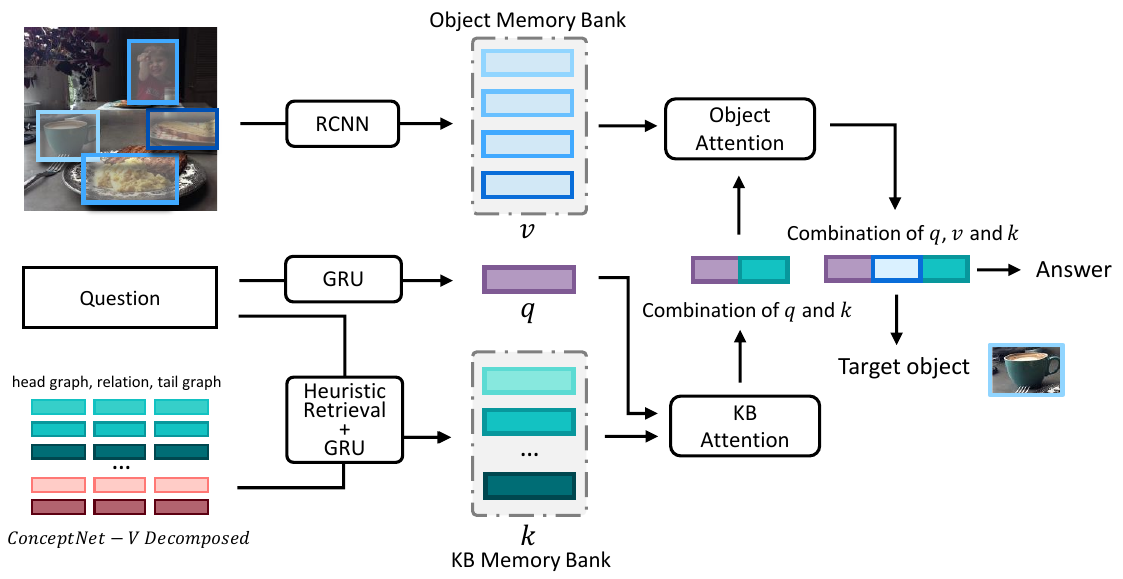}
\caption{Overall architecture of Memory-VQA model. The model encodes the image and the knowledge items into two memory banks, respectively, and then implements an iterative attention mechanism to locate relevant knowledge items and objects to answer the question.}
\label{kb_memory}
\end{figure}

\setlength{\tabcolsep}{4pt}
\begin{table*}[t]
\begin{center}
\caption{Results on the test set of the CRIC, where \textbf{Ans} indicates the answer accuracy (\%), \textbf{Grd} indicates the grounding accuracy (\%) and \textbf{Final} indicates the portion of questions on which the models both correctly generate answers and output grounding results.}
\resizebox{10.5cm}{!}{%
\begin{tabular}{lccccccc}
\toprule
\multicolumn{1}{c}{\multirow{2}{*}{Model}} & \multicolumn{2}{c}{Verify}     & \multicolumn{2}{c}{Recognize}   & \multicolumn{3}{c}{Overall} \\ \cmidrule(lr){2-8}
\multicolumn{1}{c}{}                       & \multicolumn{1}{c}{Ans} & \multicolumn{1}{c}{Grd} & \multicolumn{1}{c}{Ans} & \multicolumn{1}{c}{Grd} & \multicolumn{1}{c}{Ans} & \multicolumn{1}{c}{Grd} & \multicolumn{1}{c}{Final}                       \\ 
\noalign{\smallskip}
\midrule
\noalign{\smallskip}
%Q-Prior                      &  50.54    & -      & 15.80  & -      & 25.95  &  -    & -      \\   
%K-Prior                      &  32.13    & -      & 42.49  & -      & 39.46  &  -    & -      \\     
Q-Only                       &  68.79    & -      & 49.57  & -      & 55.18  &  -    & -      \\
I-Only                       &  48.47    & -      & 00.12  & -      & 14.24  &  -    & -      \\     
SAN                          &  75.19    & 46.45  & 59.36  & 8.38   & 63.98  & 19.50 & 17.07  \\     
Bottom-Up                    &  75.81    & 48.50  & 60.18  & 8.18   & 64.71  & 19.88 & 18.27  \\
Bottom-Up+$l_{att}$          &  73.83    & 52.90  & 57.72  & 32.06  & 62.39  & 38.10 & 29.25  \\
MAC                          &  78.71    & 52.19  & 64.91  & 23.00  & 68.91  & 31.46 & 26.19  \\     
NMN-CS                       &  79.09    & 48.69  & 64.82  & 22.60  & 68.96  & 30.17 & 25.03  \\ 
Memory-VQA                    &  76.93    & 51.36  & 62.36  & 17.99  & 66.59  & 27.67 & 23.17  \\  
Memory-VQA+$l_{att}$          &  77.44    & 61.39  & 62.64  & 44.65  & 66.93  & 49.51 & 38.87  \\  
ViLBERT                      &  86.15    & 54.21  & 71.96  & 15.97  & 76.07  & 27.06 & 23.67  \\  
ViLBERT+$l_{att}$            &  87.63    & 75.43  & 73.42  & 57.62  & 77.54  & 62.79 & 53.76  \\  

\bottomrule %  refine 
\end{tabular}%
}
\label{tab1}
\end{center}
\end{table*}
\setlength{\tabcolsep}{1.4pt}

\textbf{Memory-VQA:} We also design a simple KB-aware baseline to explicitly utilize knowledge items to answer the questions, named as Memory-VQA. This baseline follows the spirit of memory network~\cite{weston2014memory} which encodes the input materials (e.g., the knowledge items and the image in the CRIC) as memories, then uses the question to trigger an iterative attention process which allows the model to retrieve useful information to answer the question. 

The overall architecture of this model is shown in Fig.~\ref{kb_memory}. The whole model is composed of two parts. The first part realizes the feature extraction of three types of input: it implements a GRU to obtain question features $\boldsymbol{q}$, uses a Faster-RCNN to obtain image features, denoted as \emph{object memory bank} $\boldsymbol{v}$, and applies a heuristic retrieval method and a GRU to roughly select candidate knowledge items and encode them, where the output is denoted as \emph{KB memory bank} $\boldsymbol{k}$. More concretely for the heuristic KB retrieval, firstly every word in questions is used to retrieve the items in our collected ConceptNet-V-D by checking if the head or tail of a knowledge item contain such word; then a GRU selects specific relation type of items by predicting the relation type from the question. The second part implements a two-step attention to find proper knowledge items and object regions used for answering. The first step implements an attention mechanism $f_k$ ~\cite{anderson2017bottom} over the knowledge items, which first predicts attentions $\boldsymbol{a}_k$ based on the question feature $\boldsymbol{q}$ and KB memory bank $\boldsymbol{k}$ and then gives a weighted average over the KB memory bank $\boldsymbol{k}$ as the output. We then combine the output with the question embedding. The calculation of combination of question features and KB memory bank $\boldsymbol{h}_{qk}$ can be written as:
\begin{align}
\boldsymbol{h}_{qk} &= {\rm FC}(f_k(\boldsymbol{k}, \boldsymbol{q})) \odot {\rm FC}(\boldsymbol{q}), 
\end{align}
where ${\rm FC}$ denotes an FC layer and $\odot$ denotes element-wise multiplication. 
The second step uses the same architecture to calculate the attention scores over object regions $\boldsymbol{a}_v$ and gives the combined features $\boldsymbol{h}_{qkv}$ of $\boldsymbol{q}$, $\boldsymbol{k}$ and $\boldsymbol{v}$ as the output. Specifically, 
\begin{align}
\boldsymbol{h}_{qkv} &= {\rm FC}(f_v(\boldsymbol{v}, \boldsymbol{h}_{qk})) \odot {\rm FC}(\boldsymbol{h}_{qk}),
\end{align}
where $f_v$ is the attention module. Finally, an MLP predicts the answer probabilities and $\boldsymbol{a}_v$ is used to output the target object.

\textbf{Memory-VQA+$l_{att}$:} Compared to Memory-VQA, this baseline additionally applies a cross-entropy loss on attention score.

\textbf{ViLBERT:} Recently, many works~\cite{lu2019vilbert, tan2019lxmert, lu202012, chen2020uniter, li2020unicoder, li2020oscar, zhou2020unified, chen2020uniter, majumdar2020improving, zhang2021vinvl} propose powerful self-supervised learning approaches to learn the joint representations between image and language based on BERT model~\cite{devlin2019bert}. We select the ViLBERT~\cite{lu2019vilbert} pre-training on 12 vision and language datasets~\cite{lu202012} as the representative model of this branch of works. In addition, in order to output the target object, we add an attention module which is the same as the one in Memory-VQA on extracted visual features, then use attended features to output the answer.

\textbf{ViLBERT+$l_{att}$:} Compared to ViLBERT, this baseline adds a cross-entropy loss on attention score.

\begin{figure*}[t]
\centering
\includegraphics[height=7.9cm]{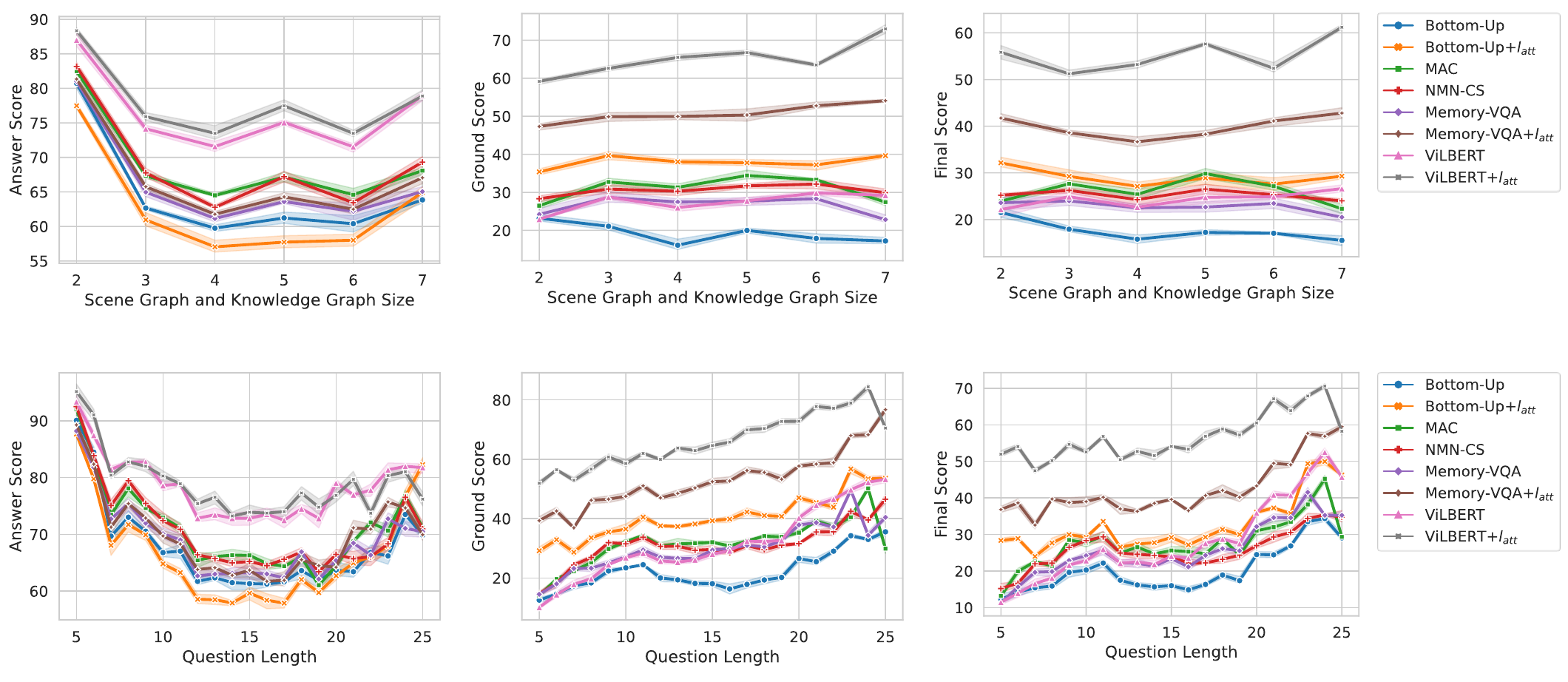}
\caption{\textbf{Top}: The performances of representative models on different sizes of question related scene and knowledge graph. \textbf{Bottom}:  The performances of models on different question lengths.}
\label{fig_exp_funsize}
\end{figure*}

\subsection{Analysis and Diagnosis}
In this section, we analyze model performances on different types of questions requiring different reasoning skills to compare existing reasoning mechanisms (Sec.~\ref{exp_overall}, Sec.~\ref{exp_qs} and ~\ref{exp_com}). Further qualitative and quantitative experiments are conducted to show the necessity of our collected additional annotations in model diagnose and training models (Sec.~\ref{exp_att} and~\ref{exp_kb}). Finally, we use modular network to investigate the main challenges of CRIC task (Sec.~\ref{exp_challenges}).

\subsubsection{Overall comparison of different types of methods}
\label{exp_overall}

The overall accuracy and the accuracy for each answer type are shown in Table~\ref{tab1}. We can see that the previous methods achieve at most 53.76\% final score. And they struggle on grounding, where the grounding accuracy of the MAC method on another VQA dataset GQA can achieve 82.2\% accuracy as reported in~\cite{hudson2019gqa} while it achieves only 31.46\% here on the CRIC. These results suggest that the CRIC cannot be solved simply by transferring the standard VQA model to such task, but requires a more delicate model to build the connections between the commonsense and images. Comparing the Bottom-Up with MAC and NMN-CS, we observe that the compositional methods achieve better results on the CRIC dataset. This could be because these models decompose the complex task into many simpler sub-tasks which is a more robust way to learn the answering skill. However, it is found that the performance gain of compositional models on CRIC is not expected as large as on CLEVR. This might be because that the cumulative error impacts the final results. For real-image and commonsense-related QA, the sub-tasks are much more difficult than corresponding sub-tasks in synthetic-images, and the new proposed commonsense-related functions still need to be solved by some sophisticated designed modules. Comparing the results of Memory-VQA and Bottom-Up, we can see that the use of external knowledge also brings significant improvement. More analysis of Memory-VQA will be presented in Sec.~\ref{exp_kb}. In addition, ViLBERT with multi-task training (i.e., ViLBERT $+l_{att}$) shows the substantial superiority compared to other models; still, the challenge is far from solved. 

%
% More delicate modules or loss functions are needed to truly ground the commonsense to the images. 

\begin{figure*}[t]
\centering
\includegraphics[width=17cm]{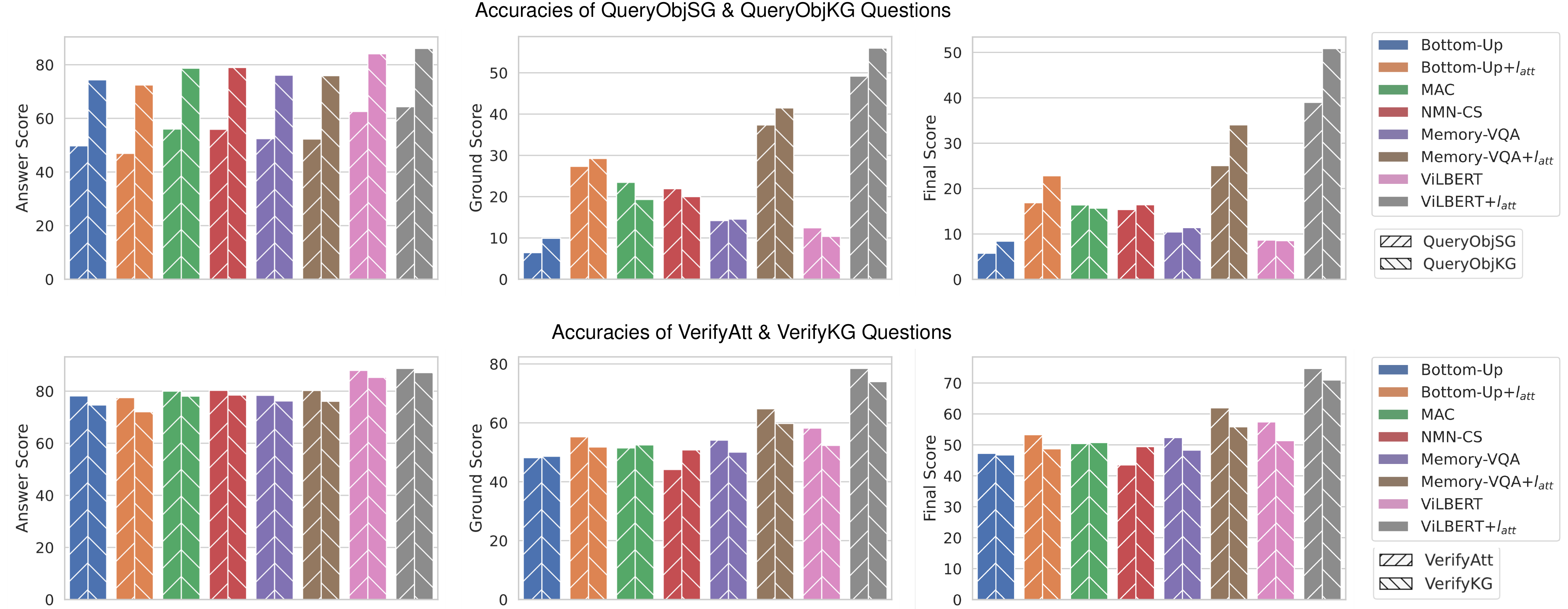}
\caption{The performances of models on different types of questions. The top row shows the scores on QueryObjSG and QueryObjKG questions. The bottom row shows the scores on VerifyAtt and VerifyKG questions. }
\label{fig_exp_kg_type}
\end{figure*}

\subsubsection{Effect of question size}
\label{exp_qs}

In this part, we evaluate different reasoning mechanisms by comparing the performances on questions in different difficulty degrees. Specifically, in Fig.~\ref{fig_exp_funsize}, we present the performances of four types of representative models on different question sizes. The question size is measured in two metrics. (1) The question size is considered as the size of question-related scene graph and knowledge graph, that is, the total number of object-attribute tuples, visual relationship triples, and knowledge items. This measure of question size represents the number of reasoning steps. (2) The question size is equal to the question length. The questions size indicates the amount of information used for object grounding.

From Fig.~\ref{fig_exp_funsize}, we can see that for \textbf{answer score}, the shapes of curves for two metrics of question size both are U-like, rather than an intuitive result that longer questions should be harder. The curves of \textbf{ground score} are also somewhat counterintuitive; with the increase of question size, the performances of models doesn't drop, but are relatively stable and even slightly better for the best two models.  The primary causes of this interesting phenomenon could be some inherent commonsense priors in questions which mirror the bias of the world. The small size questions are easier to answer because these questions usually directly query a knowledge item, so the models are likely to overuse knowledge priors to guess the answer. However, overusing priors causes the damage of correctness and robustness on grounding. Besides, the short question also provides limited information to locate the target object, while longer questions usually depict the target object from more different perspectives. Thus, though shorter questions are easier to answer, locating the object region with a concise query is still challenging. 

Such observations suggest that not only the answering long questions are difficult for current methods, for commonsense questions, grounding to images with concise commonsense query is also challenging. 

The results further demonstrate the importance of comprehensive VQA model evaluation, especially for the commonsense-related questions. The answer score and ground score play complementary roles in evaluating VQA models to better reveal their superiorities and limitations better. 

% With the increase of question length, the performances of models also doesn't drop, but even significantly better.
% TODO:
%We believe that this advantage of the CRIC makes it another dataset for the research of the commonsense VQA task.

\subsubsection{Effect of compositional commonsense questions}
\label{exp_com}

In this section, we analyze the robustness of models by comparing the performances between the compositional and the simple problem. Specifically, the questions in CRIC are divided into two groups: one group of questions directly asks the content in a knowledge item which is less compositional, e.g., \texttt{QueryObjKG}, \texttt{VerifyKG}; another group requires an indirect use of knowledge items which are more compositional, e.g., \texttt{QueryObjSG}, \texttt{VerifyAtt}. In the Fig.~\ref{fig_exp_kg_type}, we compare the performances of models on these two groups of questions, \texttt{QueryObjSG} vs. \texttt{QueryObjKG}, and \texttt{VerifyKG} vs. \texttt{VerifySG}. From the results of \textbf{answer score} of query-type questions (top left figure), we can see a large gap between \texttt{QueryObjSG} and \texttt{QueryObjKG}. While for \textbf{ground score} (top middle figure), there is only a small gap between the two types of questions. In other words, under a similar grounding ability, it is more difficult to answer questions which indirectly query the commonsense. 
This demonstrates that the compositional questions effectively evaluate whether the model really understands the vision and commonsense. 
Besides, the performances of a model are very close on \texttt{VerifyAtt} and \texttt{VerifyKG} questions (bottom figures). This may be because the main function of knowledge prior is to reduce the number of candidate answers, while this function is invalid in answering verify-type questions.

These results also reflect that increasing the grounding performances is one of the most important direction for current methods. It limits the performance of compositional questions.

\subsubsection{Effect of attention supervision}
\label{exp_att}
In this section, we show the importance of our collected additional attention annotations in model training and evaluation. In Table~\ref{tab1}, we present three sets of models, Bottom-Up, Memory-VQA, ViLBERT, and their corresponding versions with attention supervision. We can see that adding attention supervision brings significant improvements in ground scores and final scores for all three models and slight improvements in answer scores for Memory-VQA and ViLBERT. This shows that even if the model has an explicit attention module, it is still difficult for the model to learn a robust object localization spontaneously. Especially for questions whose answers come from knowledge items, in the top middle of Fig.~\ref{fig_exp_kg_type}, we can see that the models achieve larger improvement on \texttt{QueryObjKG} questions after adding attention supervision. 
Besides, it is found that the stronger the model, the more noticeable this phenomenon. ViLBERT itself is a very outstanding model in object localization proven in many other tasks~\cite{lu202012}, but it is difficult to give full play to this advantage without adding appropriate supervision information. 

These results suggest that our attention annotations are also critical for evaluation and we may need to utilize the attention annotations to increase the model's robustness and allow the model to output meaningful intermediate results. And 

\subsubsection{Effect of explicit use of knowledge items}
\label{exp_kb}

The main difference between Bottom-Up and Memory-VQA is that the latter has an additional branch that explicitly utilizes the knowledge items to answer the question. From the results of these two types of models in Table~\ref{tab1}, we can see that the explicit use of commonsense knowledge not only brings an improvement in answering, but also a significant improvement in grounding performance ($>$ 9\% absolute improvement). This is an interesting phenomenon: what we provide to the model is actually more clear knowledge prior information, but this information does not exacerbate the phenomenon of overfitting on priors (i.e., achieving higher answer score and lower ground score), instead, it helps the model gain better robustness and learn more meaningful intermediate result (i.e., achieving an obvious improvement in grounding). 

%This also demonstrate the 

\begin{figure}[t]
\centering
\includegraphics[height=17.0cm]{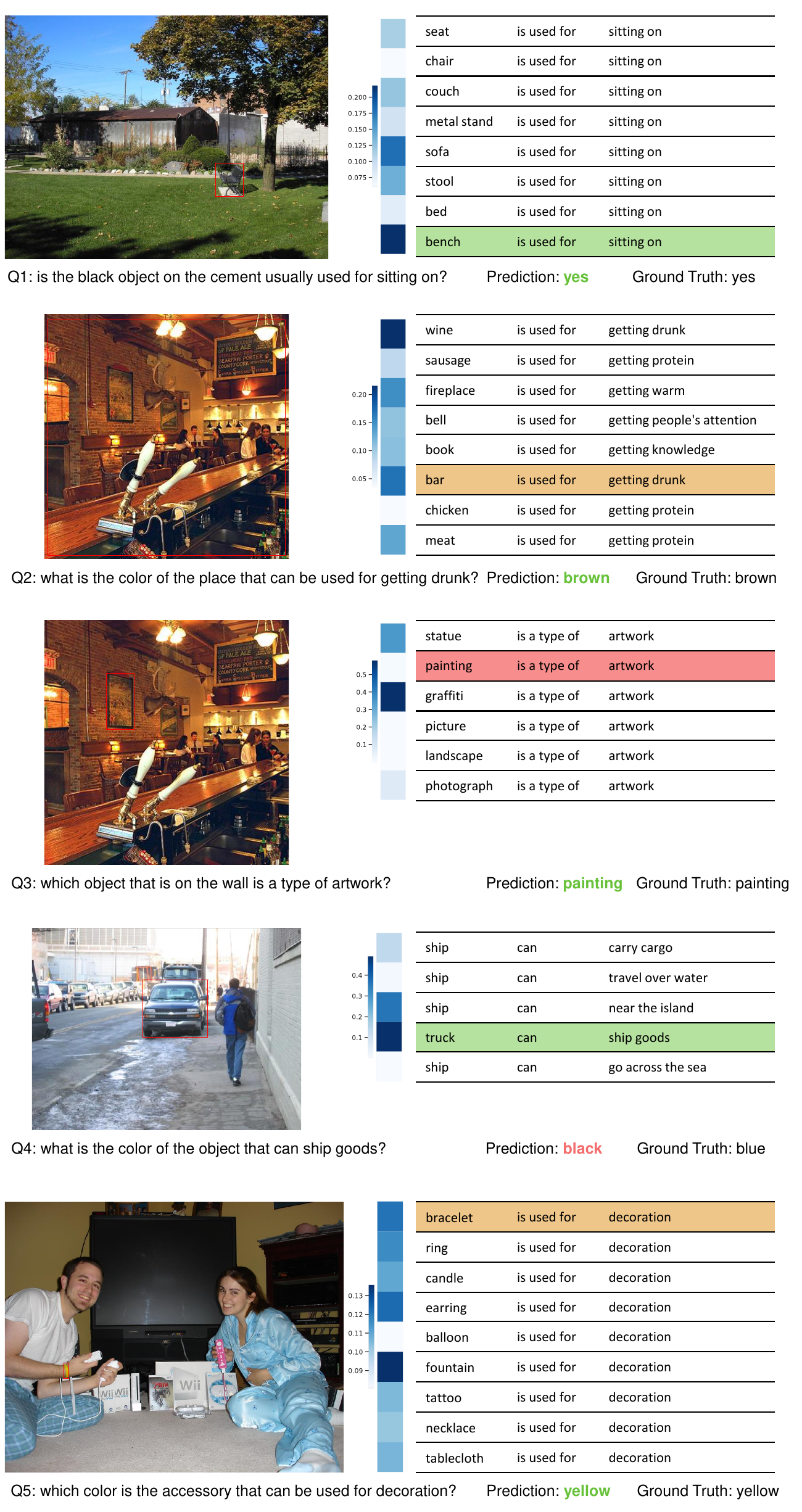}
\caption{Visualization of Memory-VQA+$l_{ans}$'s attention scores on knowledge items. The highlighted knowledge items are the items used for generating their question. The red bounding boxes are predicted target objects.}
\label{fig_mem_att}
\end{figure}

In the Fig.~\ref{fig_mem_att}, we show the visualized attention scores of Memory-VQA$+l_{att}$ over knowledge items. We can see that the model usually attends on the relevant knowledge item with a higher score (the items in green or orange background). But, when many object categories meet the commonsense requirement of a question, it is still difficult to accurately locate the most relevant knowledge item, e.g., \textbf{Q3}. In other words, it is crucial for solving the CRIC task to design models to align knowledge and visual content locally. Besides, we also find that the distribution of the model's attention scores is relatively uniform. This is consistent with the intuition that the model not only needs the knowledge item related to the target object, but also other knowledge items that help exclude the wrong objects. 

The above experiments show that explicitly utilizing knowledge items is very effective. Only implementing the most straightforward mechanism to explicitly use knowledge items shows obvious improvements. We believe that further exploring some techniques about jointly representing vision and knowledge is crucial, e.g., how to align the commonsense and vision, how to represent knowledge items, and how to use multiple knowledge items simultaneously.

\subsubsection{Challenging subtasks of the CRIC}
\label{exp_challenges}

To better display the CRIC dataset's challenges, we conduct an additional experiment that tests the performance of each module in NMN-CS which can access ground-truth attention inputs and text inputs to purely evaluate the difficulty of sub-tasks (denoted as NMN-CS-GT). More specifically, the outputted attention scores of grounding-related modules are passed through a sigmoid layer to determine if attending on the corresponding object candidate or not. In addition, the grounding modules are trained by ground-truth attention outputs with the weighted binary cross-entropy loss, where the weighted loss is to tackle the problem that attended objects and background objects are highly imbalanced. And the recognition module is trained by binary cross-entropy loss. 

\begin{figure*}[t]
\centering
\includegraphics[height=16.0cm]{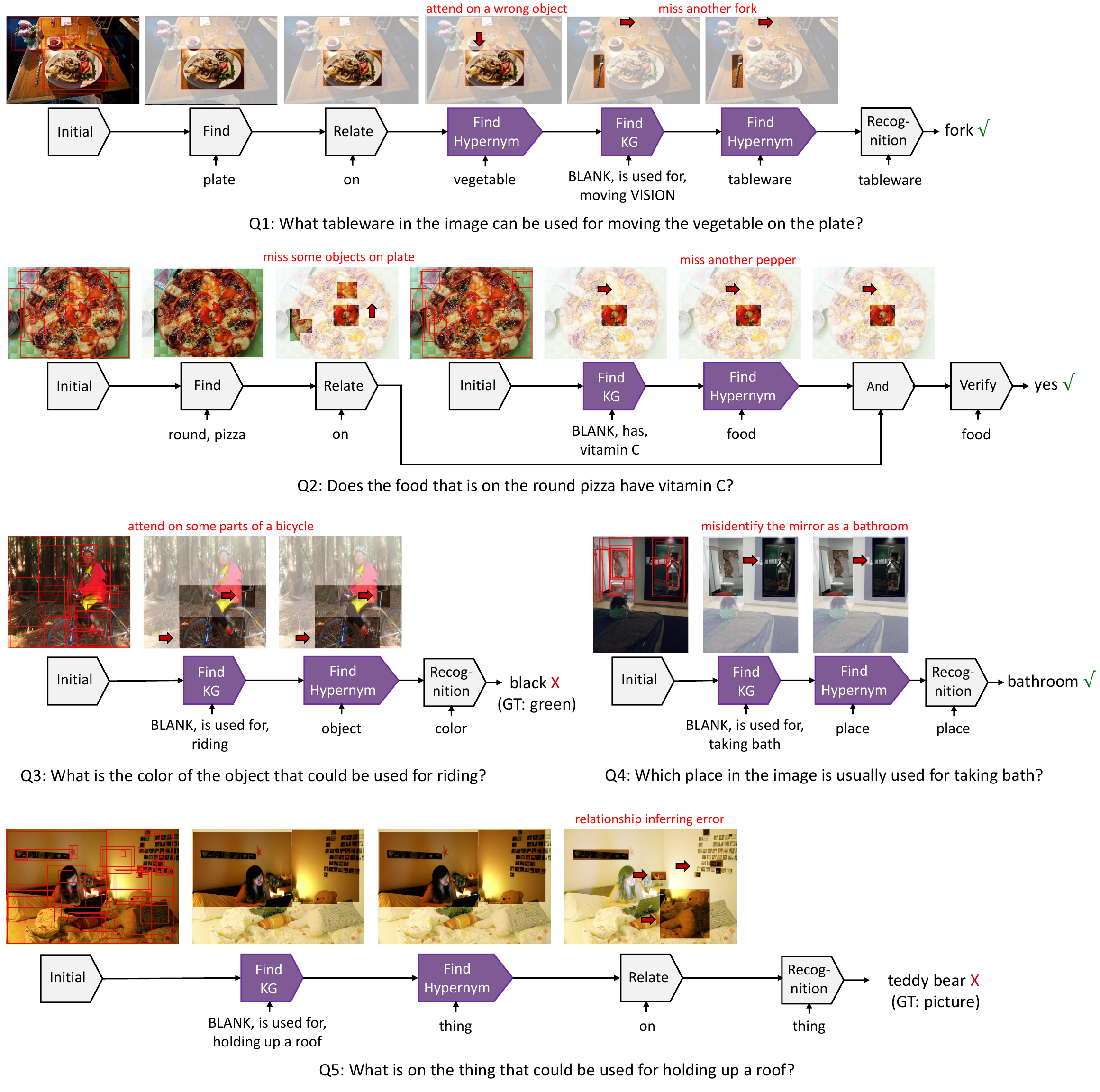}
\caption{Question answering examples of NMN-CS-GT which uses ground-truth text inputs. Though the model correctly answers the questions, the model still makes some mistakes in the intermediate modules. Precisely accomplishing each sub-task is still challenging.}
\label{fig7}
\end{figure*}

For program prediction, the accuracy of function name prediction is 99.25\%. This indicates that understanding the question is relatively easy. For grounding-related modules, we define the ground score as the IoU score of the output object set $O_{predict}$ and the ground-truth object set $O_{gt}$, 
\begin{equation}
score = \frac{|O_{predict} \cap O_{gt}|}{|O_{predict} \cup O_{gt}|},
\end{equation}
where $|.|$ indicates the number of elements in a set. The IoU score is a strict metric because a slight difference with $O_{gt}$ will cause a relatively low score when $|O_{gt}|$ is small, so this metric can better expose the flaw of the grounding model.

\setlength{\tabcolsep}{4pt}
\begin{table}[t]
\begin{center}
\caption{The performance of each module in NMN-CS-GT. The scores of grounding-related module are IoU scores of predicted object set and ground-truth object set. The score of Recognition module is prediction accuracy.}
\label{fun_acc}
\resizebox{8.50cm}{!}{%
\begin{tabular}{lccccc}
\toprule
      & Find  & Relate & Find\_Hypernym & Find\_KG  & Recognition \\ \midrule
Score & 40.56 & 18.03  & 47.61          & 16.72     & 55.34        \\ \bottomrule
\end{tabular}%
}
\end{center}
\end{table}
\setlength{\tabcolsep}{1.4pt}

From the results in Table~\ref{fun_acc}, we can identify that \texttt{Find} and \texttt{Find\_Hypernym} tasks are relatively easy, but it is still far from satisfactory. In addition, though the score of \texttt{Find\_Hypernym} is higher than Find, it doesn't mean that \texttt{Find\_Hypernym} is much easier. The \texttt{Find\_Hypernym} sometimes requires the model to locate objects belonging to ``object'' or ``thing'', which is relatively easy, while correctly finding objects belonging to a category, such as ``furniture'' or ``vegetable'', is still challenging. Moreover, \texttt{Find\_KG} related tasks are relatively more difficult than others. Along with the phenomenon in Table~\ref{tab1} that many models achieve high accuracy on answering, it shows that it is easy to understand the commonsense in semantic-level. Still, it isn't easy to learn to ground the commonsense knowledge into the images. It might need to propose a more advanced model to encode the commonsense items to better align with visual objects. For the \texttt{Recognition} task, the score is the accuracy of the output. We can see that the Recognition task is also relatively difficult for current models. This might be because, the visual genome dataset involves a large number of visual concepts, including thousands of object categories and attributes with diverse semantics. 

In Fig.~\ref{fig7}, we show the qualitative results of the NMN-CS-GT, where modules are separately trained with ground-truth function output. It can be seen that even though the whole model predicts the correct answer, it is still challenging for each module to provide precise intermediate attention results. For example, the model correctly predict that \emph{fork} can be used for moving the vegetable for \textbf{Q1}, while it doesn't correctly find the vegetable and all forks in the image. 

In brief, many sub-tasks of CRIC is difficult and the cumulative error is the main bottle neck restricting the multi-step reasoning performance of modular networks. We may need to propose new representations to make a significant improvement of each module, or present a new framework of modular network to avoid the cumulative error. Besides, it also shows that our provided rich ground-truth annotations can assist in diagnosing and improving the robustness of future models.

\section{Discussions}
\label{discussion}

\hspace{\parindent}\textbf{Comparison with FVQA, GQA, and OK-VQA.}
The CRIC extends the VQA task along two directions: towards multi-hop reasoning and understanding of non-visual knowledge of multiple objects, as illustrated in Fig.~\ref{fig1}. Here, we introduce more details about the key features of the CRIC that differentiate it from existing ones, e.g., FVQA, GQA, and OK-VQA.

1) The CRIC introduces new types of compositional commonsense-related questions, e.g. QueryObjSG, VerifyAtt, etc., to evaluate some unique capabilities. The questions in GQA mainly measure the understanding of visual contents (no external knowledge sources are used to generate questions). The pioneering commonsense VQA work, FVQA, primarily focuses on QueryObjKG type questions with limited spatial reasoning. The questions of OK-VQA mainly evaluate the breadth of knowledge of the models. It requires the model to crawl useful information from the internet.
In comparison, our new types of questions mainly aim to investigate whether the model can truly ground the commonsense into the visual world, as illustrated in Sec.~\ref{exp_com}.

2) As far as we know, our dataset is the first large scale commonsense VQA datasets where attention results, the program for answering questions and question-related scene graph and knowledge graph are all available. Because of that, we can provide a more comprehensive platform to support the evaluation of various types of methods. Besides, in Sec.~\ref{exp_att} and Sec.~\ref{exp_kb}, we also show that different types of annotations facilitate developing more robust commonsense VQA models. Multiple types of annotations also facilitate different types of models to be merged. 

\textbf{Open Issues of Auto-Generated Datasets.}  As another new automatically-generated dataset, our CRIC inherently has the following advantages: 1) Ease the risk of overusing priors. As stated above, the CRIC addresses this issue in each vital stage during construction. 2) Provide rich annotations for detailed evaluation and diagnosis. 3) Easy to measure the complexity of questions. The number and the types of sub-tasks involved in a question can serve as useful and direct indicators to automatically assess its complexity for better diagnosis. 4) Easy to extend the dataset on new images or knowledge items. While embracing such favorite features at a low cost of human labor, like other auto-generated datasets (e.g., CLEVR, GQA), our CRIC faces the challenge of maintaining questions' naturalness. To tackle this issue, a certain level of human interference, like manually rephrasing the questions, would be necessary in the future.

\section{Conclusion}

This paper introduces the CRIC dataset that evaluates VQA systems on answering questions requiring compositional reasoning on the vision and commonsense. To build this dataset, we first propose a new Knowledge Graph format for easily aligning knowledge items to visual entities and depicting the commonsense relations between objects. Then, we propose an efficient method to generate numerous QA pairs and rich annotations automatically. Our generation method has better scalability and requires lower cost, easing the difficulty of building a complex VQA dataset. 

Further experiments analyze the current four representative types of models. The results demonstrate our annotations' effectiveness on both comprehensive evaluation and enhancing the models' performances and robustness. The CRIC also brings new challenges for representation and reasoning of vision, question, and knowledge, e.g., how to design a model to capture the joint of graphs' global information in two modalities; how to conduct multi-hop reasoning on these two graphs explicitly; how to uniformly represent the commonsense and vision to better ground the commonsense. And various types of annotations will help researchers integrate the ideas of multiple types of models or propose a new unified framework to solve these challenges. 
For example, redesign the BERT as a modular network, equip modular networks with the ability of explicitly using the commonsense, or propose a pre-training model processing vision, language, and knowledge items simultaneously.
In brief, we hope the CRIC can help drive the research of more transparent and robust models for designing more advanced AI agent reasoning on the vision and commonsense.

\bibliographystyle{IEEEtran}
\bibliography{egbib}

\ifCLASSOPTIONcaptionsoff
  \newpage
\fi

\clearpage

\begin{onecolumn}
\centerline  {\Large \textbf{Appendix}}
\section*{Overview}
In the appendix, we provide more details of CRIC dataset and model NMN-CS:

\noindent
\ref{app_fun} - Function Definition of CRIC

\noindent
\ref{sup_nmn} - Details of NMN-CS

\noindent
\ref{sup_c} - Some QA samples in CRIC

\section{Function Definition of CRIC}
\label{app_fun}
In this section, we introduce the 10 basic functions that our dataset aims to evaluate. These functions operate on some values that are indicated in a question or generated by some neural modules and output an object list or a concept.

\textbf{Inputs of functions.} These basic functions have two types of inputs. The first one is the text input that is indicated in a question:
\begin{itemize} %[itemsep= -2 pt,topsep = -1 pt]
%\vspace{-0.1cm}
	\item \texttt{object}: An object name, e.g., \emph{dog, double decker}.
%\vspace{-0.1cm}
	\item \texttt{attribute}: An attribute name, e.g., \emph{blue, open}.
%\vspace{-0.1cm}
 	\item \texttt{predicate}: A predicate name, e.g., \emph{on, holding}.
%\vspace{-0.1cm}
 	\item \texttt{type}: A category name that indicates recognizing one type of concepts, e.g., \emph{color, object, animal}. 
%\vspace{-0.1cm}	
	\item \texttt{KG\_Query}: A commonsense query indicates a knowledge item, where some elements are replaced with BLANK or VISION for specific functions, e.g., $<$\emph{BLANK, Can, climb the VISION}$>$, which means finding objects that can climb a specified object (e.g., a tree) in an image.
%\vspace{-0.1cm}
	\item \texttt{hypernym}: A category name that indicates finding objects belonging to one category, e.g., \emph{animal, furniture}. 
\end{itemize}

Another type of input is a vector generated by some other modules:
\begin{itemize}[itemsep= -2 pt,topsep = -1 pt]
\vspace{-0.1cm}
\item \emph{objects}: A set of objects (could contain zero, one or multiple objects) in an image.
\end{itemize}

% (only show questions of GQA on the middle row)

\textbf{Outputs of functions.} Our functions have two types of outputs:
\begin{itemize} %[itemsep= -2 pt,topsep = -1 pt]
%\vspace{-0.1cm}
\item \emph{objects}: A set of objects in a given image.
\item \emph{concept}: A concept that could be the name of a visual concept (object, attribute, scene, etc.) or a boolean value (indicates yes or no).  
\end{itemize}

\textbf{Basic Functions.} In this part, we introduce 7 visual basic functions (the other 3 commonsense functions have been illustrated in our main paper in Sec.3 \textbf{Function Definition}).
\begin{itemize} %[itemsep= -1 pt,topsep = 0 pt]
\item Find: Given a set of objects, filter the objects by the object name or the attribute name or both two, e.g., find ``cat'', find ``black'', find ``black cat''.  

\item Relate: Return all objects in the image that have the specified relation \texttt{predicate} to the input objects, where input objects are the ``subject'', output objects are the ``object''. For example, find all objects that the man (``subject'') is holding (``predicate'').

\item Relate Reverse: Return all objects in the image that have the specified relation \texttt{predicate} to the input objects,  where input objects are the ``object'', output objects are the ``subject''. For example, find all objects that are on (``predicate'') the table (``object'').

\item Recognition: Recognize the concept in the \emph{objects} among one \texttt{type} of concepts, e.g., recognize the color in one image region. 
%Note that, we only apply Recognition function when the \emph{objects} contain only one object.

\item And: Return the intersection of two sets of objects. 
\item Verify: Given a set of objects, output \emph{yes} if the set is non-empty and \emph{no} if it is empty.
\item Initial: Output the set of all objects in the image. 
%\item Ground KG: Given a commonsense query \texttt{relation}, \texttt{tail}, find all the objects that satisfied the query $<$object, \texttt{relation}, \texttt{tail}$>$, e.g. find the region contains the object that $<$ \_ , \emph{can be used for}, \emph{drying wet hair}$>$.
%\item Query KG: Infer the \texttt{tail} for given a concept (indicates the \texttt{head}) and the \texttt{relation}, where $<$\texttt{head}, \texttt{relation}, \texttt{tail}$>$ is a correct knowledge triplet. 
%\item Verify KG: Judge the correctness of the statement in the question,  e.g. if the object $<$ \_ , \emph{can be}, \emph{eaten}$>$.
\end{itemize}

\begin{table*}[hp]
\caption{The details of visual modules in NMN-CS. Each function takes text inputs $\bold{t}$ indicated in the question (i.e., the attended question feature), some attention inputs, such as $\bold{a}$,  $\bold{a_1}$, and $\bold{a_2}$, generated by some other modules, and features of all objects $\bold{v}$ as inputs, then achieves corresponding function and outputs an attention map $\bold{y}_a$ (shorted as ``att'') or a probability vector $\bold{y}_c$ over all candidate answers. The operator $\odot$ is element-wise multiplication, ${\rm sum}$ is summing the results over spatial dimensions, and $\bold{e}_N$ is an N dimensional (the number of objects in the image) vector where all elements are 1.}
\centering
\resizebox{\textwidth}{!}{%
\begin{tabular}{lccccc}
\toprule
Module                 & text inputs & attention inputs & output &  Implementation details \\ \midrule
Find                   & $\textbf{t}$ & $\bold{a}$& att  & $\bold{y}_a = {\rm sigmoid}({\rm FC}({\rm FC}(\bold{a} \odot \bold{v}) \odot {\rm FC}(\bold{t}))$         \\
Relate/Relate\_Reverse & $\textbf{t}$       &$\bold{a}$& att  & $\bold{y}_a = {\rm sigmoid}({\rm FC}({\rm FC}(\bold{v}) \odot {\rm FC (sum}({}\bold{a} \odot \bold{v})) \odot {\rm FC}(\bold{t})))$ \\
Recognition            & $\textbf{t}$       & $\bold{a}$ & concept & $\bold{y}_c = {\rm sigmoid}({\rm FC}({\rm FC}({\rm sum}(\bold{a} \odot \bold{v})) \odot {\rm FC}(\bold{t})))$                                     \\
And                    & (none)      & $\bold{a}_1$, $\bold{a}_2$ & att       & $\bold{y}_a = {\rm sigmoid}({\rm FC}({\rm FC}(\bold{a}_1) \odot {\rm FC}(\bold{a}_2)))$        \\
Verify                 & (none) & $\bold{a}$ & concept  & $\bold{y}_c = {\rm softmax}({\rm FC}(\bold{a}))$ \\
Initial                & (none) & (none)     & att      & $\bold{y}_a = \bold{e}_N$ \\ \bottomrule
%Find\_KG/\_Reverse             & $t_r$, $t_t$   &  $\bold{a}$      &  att      & $\bold{y}_a = {\rm sigmoid}({\rm FC}({\rm FC}(\bold{a} \odot \bold{v}) \odot {\rm FC}(\bold{t}_t - \bold{t}_r)))$        \\
%Hypernym\_KG              & $t_r$      &  $c$      &   concept     & $y_c = {\rm argmax}({\rm sigmoid}({\rm FC}(\bold{c} + \bold{t}_r)))$        \\  
\end{tabular}%
}
\label{nmn}
%\vspace{-0.4cm}
\end{table*}

\section{Details of NMN-CS}

\label{sup_nmn}
In Sec.4 of the body, we illustrate the commonsense-related modules of NMN-CS. In Table~\ref{nmn}, we show the details of visual modules in NMN-CS. A visual module is a function $y = f(\bold{a}_1, ..., \bold{a}_n, \bold{v}, \bold{t})$ that takes $n$ ($n \in \{0,1,2\}$ in our model, and $n=0$ indicates that no $\bold{a}_i$ is inputted into the function) tensors ($\bold{a}_1, ...., \bold{a}_n$) generated from other neural modules, image features $\bold{v}$ and text input feature $\bold{t}$ extracted from the question (an attended question features generated by program prediction module) as inputs, and outputs a tensor $\bold{y}$ which is either an attention map $\bold{a}$ over image regions or the probability $\bold{c}$ over all possible answers.

\section{Some QA samples in CRIC}
\label{sup_c}
In Fig.~\ref{sup_samples}, we show more QA samples in the CRIC dataset and some questions from GQA and VQA v2, which share the same images with the CRIC to present the differences between the datasets.

\begin{figure*}[tp]
\centering
\includegraphics[height=22.3cm]{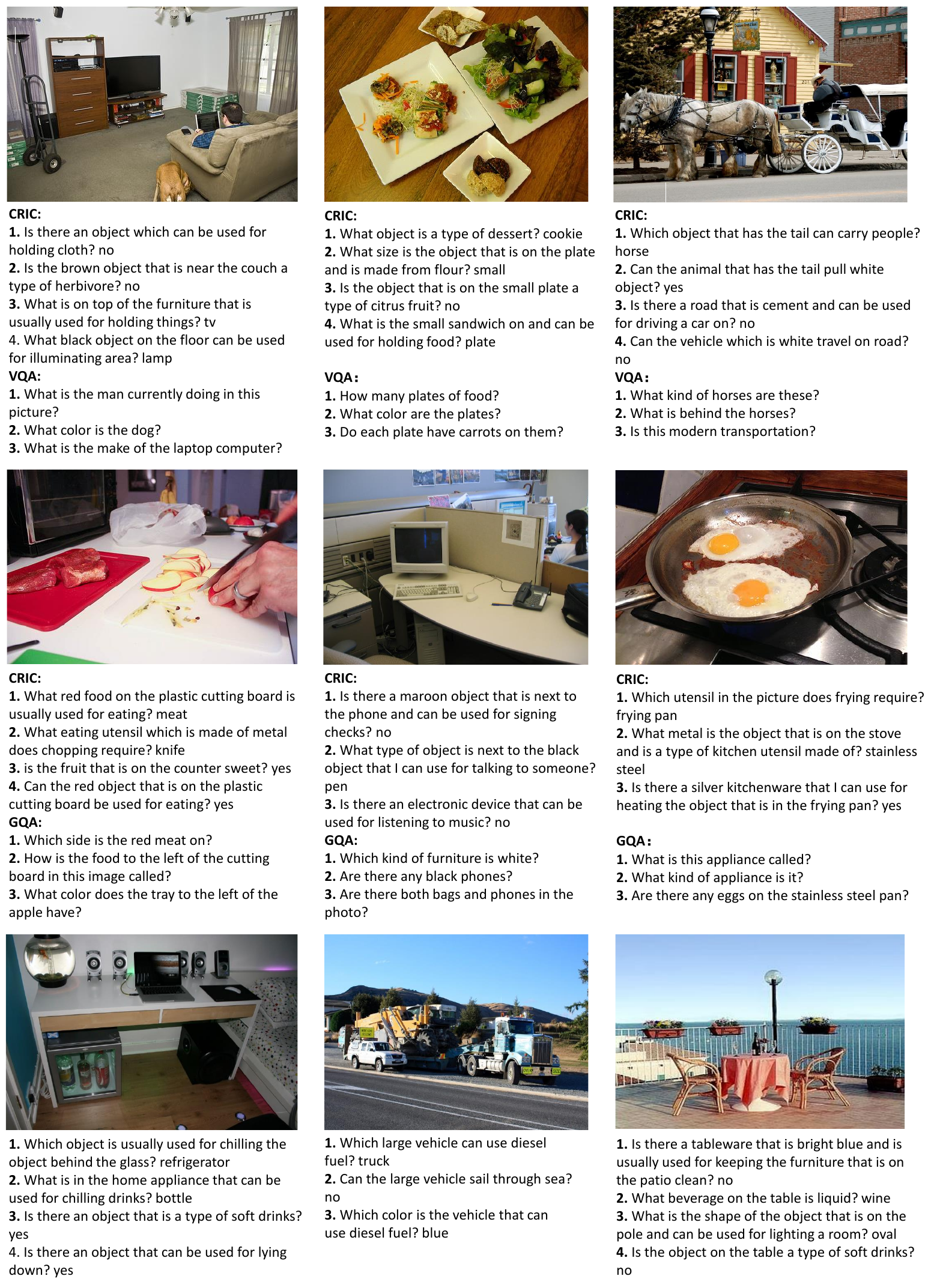}
\caption{The top one row displays the COCO images in Visual Genome and the corresponding questions from CRIC and VQA~\cite{goyal2016making}. The middle and bottom rows display the Flickr images in Visual Genome and the corresponding questions from CRIC and GQA~\cite{hudson2019gqa}. VQA v2~\cite{goyal2016making} mainly focuses on questions about querying visual facts with less compositionality. More recent GQA~\cite{hudson2019gqa} mainly evaluates the complex compositional reasoning on visual facts. While CRIC contains compositional questions requiring reasoning on visual and commonsense.}
\label{sup_samples}
\end{figure*}

% \bibliographystyle{IEEEtran}
% \bibliography{egbib}

%The authors would like to thank...

\end{onecolumn}

\end{document}